\documentclass[sigconf,screen]{acmart}

\usepackage{xcolor}
\usepackage{graphicx}
\usepackage{booktabs}
\usepackage{tabularx}
\usepackage{amsmath}
\usepackage{makecell}
\usepackage{subcaption}
\usepackage{float}
\usepackage{multirow}
\usepackage{array}
\usepackage{wrapfig}

\AtBeginDocument{%
  }

\setcopyright{acmlicensed}
\copyrightyear{2025}
\acmYear{2025}
\acmDOI{10.1145/3709025.3712219}

\acmConference[CS\&Law ’25]{4th ACM Symposium on Computer Science and Law}{March 25-27, 2025}{Munich, Germany}
\acmISBN{979-8-4007-1421-4/2025/03}

\begin{document}

\title{A Reasoning-Focused Legal Retrieval Benchmark}

\author{Lucia Zheng}
\authornote{Equal contribution.}
\affiliation{%
  \institution{Stanford University}
  \city{Stanford}
  \state{California}
  \country{USA}
}
\email{zlucia@stanford.edu}

\author{Neel Guha}
\authornotemark[1]
\affiliation{%
  \institution{Stanford University}
  \city{Stanford}
  \state{California}
  \country{USA}
}

\author{Javokhir Arifov}
\affiliation{%
  \institution{Stanford University}
  \city{Stanford}
  \state{California}
  \country{USA}
}

\author{Sarah Zhang}
\affiliation{%
  \institution{Stanford University}
  \city{Stanford}
  \state{California}
  \country{USA}
}

\author{Michal Skreta}
\affiliation{%
  \institution{Stanford University}
  \city{Stanford}
  \state{California}
  \country{USA}
}

\author{Christopher D. Manning}
\affiliation{%
  \institution{Stanford University}
  \city{Stanford}
  \state{California}
  \country{USA}
}

\author{Peter Henderson}
\authornote{Equal advising.}
\affiliation{%
  \institution{Princeton University}
  \city{Princeton}
  \state{New Jersey}
  \country{USA}
}

\author{Daniel E. Ho}
\authornotemark[2]
\affiliation{%
  \institution{Stanford University}
  \city{Stanford}
  \state{California}
  \country{USA}
}

\renewcommand{\shortauthors}{Zheng et al.}

\begin{abstract}
    As the legal community increasingly examines the use of large language models (LLMs) for various legal applications, legal AI developers have turned to retrieval-augmented LLMs (``RAG'' systems) to improve system performance and robustness. An obstacle to the development of specialized RAG systems is the lack of realistic legal RAG benchmarks which capture the complexity of both legal retrieval and downstream legal question-answering. To address this, we introduce two novel legal RAG benchmarks: Bar Exam QA and Housing Statute QA. Our tasks correspond to real-world legal research tasks, and were produced through annotation processes which resemble legal research. We describe the construction of these benchmarks and the performance of existing retriever pipelines. Our results suggest that legal RAG remains a challenging application, thus motivating future research.
\end{abstract}

\ccsdesc[500]{Computing methodologies~Natural language processing}
\ccsdesc[500]{Applied computing~Law}

\keywords{retrieval, reasoning, benchmark, dataset}


\maketitle

\section{Introduction}
There is significant excitement towards using large language model (LLM) tools to improve the quality and cost of legal services~\cite{ma2024generative}. Already, lawyers across the world have begun incorporating LLMs into legal practice, and applying them towards a range of tasks: answering questions about the law in various jurisdictions, identifying potential legal issues in client cases, drafting agreements, and more~\cite{bommasani2021opportunities}. 

Applying LLMs towards legal tasks requires resolving distinctive challenges posed by the legal domain. For instance, legal tasks are often fact-intensive~\cite{guha2024legalbench} and LLMs have a tendency to produce factually-ungrounded statements (``hallucinations'')~\cite{dahl2024large}. The law is also constantly changing---through new statues or judicial opinions---and the knowledge contained in LLM parameters is static~\cite{NEURIPS2021_f5bf0ba0}. To address these challenges, legal AI developers have begun deploying ``RAG'' systems, where  LLMs are augmented with retrievers over corpora of case law, statutes, and other legal documents~\cite{magesh2024hallucinationfree}. Given a lawyer query, the retriever fetches documents from the corpora relevant to the query, and requires the LLM to answer the query with respect to the retrieved documents.

However, a significant bottleneck in the development of legal RAG systems is the lack of realistic English legal open-domain question answer benchmarks. In particular, existing benchmarks  suffer from one or more of the following weaknesses.
\begin{enumerate}
    \item First, they fail to represent tasks where where the query and relevant document have little lexical overlap, and identifying the relevant document requires multi-hop or analogical reasoning. In practice this setting is ubiquitous. Producing the legal cases relevant to a client's factual circumstances, for instance, requires extracting higher-order legal issues and identifying other cases which present those issues---even if the specific factual descriptions are quite different.
    \item Second, existing benchmarks are often exclusively retrieval benchmarks, and do not contain paired question-answers to evaluate downstream reasoning based on the retrieved information~\citep{coliee2024, hou2024clerc, tyss2024ecthr, mahari2023lepard, chalkidis-etal-2021-regulatory}. As a result, they do not capture the downstream impacts of improvements in retrievers.
    \item Finally, benchmarks rely on query-document distributions extracted from datasets built for other purposes, where queries do not correspond to the types of questions lawyers might actually ask~\cite{saad2024benchmarking, hou2024clerc, coliee2024}.
\end{enumerate}

To address this gap, we introduce two new benchmark datasets for evaluating retrieval-augmented LLMs: Bar Exam QA and Housing Statute QA. These datasets address the deficiencies discussed above. In Bar Exam QA,  queries correspond to reasoning-intensive  Bar Exam hypothetical fact-patterns, and documents correspond to judicial opinion passages necessary for answering the hypothetical. In Housing Statute QA, queries correspond to practically useful questions about housing law, and documents correspond to statutes from different jurisdictions. Our datasets provide \textasciitilde10K labeled, paired query, gold passage, answer examples for training and evaluating language models on legal retrieval and retrieval-augmented downstream QA tasks. The gold passages for these datasets are hand-annotated and validated by law students and researchers, through annotation processes modeled off of a lawyer's legal research process.

Concretely, our work makes three contributions. First, we describe the construction of these datasets (Section \ref{sec:datasets}) and compare them to existing benchmarks (Section \ref{sec:task_complexity_analysis}). We show that relative to existing benchmarks, ours captures query-document distributions where the lexical similarity between the query and document is low. Second, we benchmark existing state-of-the-art retrieval pipelines on these datasets and find that because of the low lexical similarity, common retrieval methods like BM25 struggle (Section \ref{sec:baselines}). Third, we present results of a simple heuristic to verify that our benchmarks accurately measure improved legal reasoning in retrieval (Section \ref{sec:expansion}).  Specifically, we describe a law-inspired query expansion strategy with generative reasoning roll-outs. We find that this approach improves performance on our datasets.

Our work suggests that developers of retrieval-augmented legal LLM products may need to go further than simple retrievers to improve the performance of their approaches. In particular, they may need to ensure that retrievers can also be legal reasoners too, either through query expansion or increased embedding model capacities.

\section{Related Works}
\subsection{Open-Domain QA Datasets}
BEIR \cite{thakur2021beir} is a widely used information retrieval (IR) benchmark, which consists of 18 datasets across 9 task types. Our datasets are most similar to the question answering tasks in BEIR, Natural Questions (NQ) \cite{kwiatkowski2019nq}, HotpotQA \cite{yang2018hotpotqa}, and FiQA-2018 \cite{10.1145/3184558.3192301}. For this style of task, the retrieved passage is used as context to help the model on downstream question answering. For NQ and HotpotQA, the answer is an extractive span of the context passage. For Bar Exam QA and Housing Statute QA,  the answer is in a multiple-choice format.

Past works have also discussed several limitations of existing general IR benchmarks: a skew towards web/search-engine style retrieval tasks, tasks with low lexical and syntactic variance between queries and gold passages, tasks with short query lengths, and propose new datasets to address these challenges \cite{JoshiTriviaQA2017, thakur2021beir}. 
Most similar to our tasks, BIRCO \cite{wang2024birco} and BRIGHT \cite{su2024bright} introduce new benchmark IR tasks with more complex task objectives and that require greater reasoning capabilities to solve. Compared to BEIR \cite{thakur2021beir}, these tasks have significantly lower lexical similarities between queries and gold passages and longer query lengths.
BIRCO and BRIGHT include reasoning tasks in the natural sciences, computer science, and theorem-based mathematics, but neither contains legal reasoning tasks that share similar types of deductive reasoning processes to mathematical reasoning tasks. 

\subsection{Legal Information Retrieval Datasets}
Early work on retrieval of statutory law focus on building systems using lexical matching and extensive annotation of semantic features \cite{grabmiar2015luima, vsavelka2022legal}.

More recently, several works release legal retrieval datasets constructed by leveraging case document structure/metadata to link a citing context in a new case (query) to precedential (prior) cases cited to support arguments made in the citing context (gold passage) \cite{coliee2024, hou2024clerc, mahari2023lepard, chalkidis-etal-2021-regulatory}. Though this automatic extraction approach enables the collection of large-scale datasets,  the citing contexts often summarize the high-level rule from the cited case relevant to the argument to justify its citation.\footnote{The ECtHR-PCR dataset \cite{tyss2024ecthr} is an exception. The authors leverage additional case document structure to separate facts from the argument in the citing case context when constructing queries.} As a result, we find the lexical similarity of the query and the gold passage is often quite high and comparable to ODQA datasets for these legal IR datasets (Section \ref{sec:task_complexity_analysis})

Additionally, since the queries are extracted directly from case opinions written by judges, they often do not reflect the natural distribution of user question-style queries that might be asked by a person seeking legal information or a lawyer conducting legal research. To our knowledge, there are few English-language legal IR datasets with natural question-style queries and expert gold passage annotations. Existing datasets \cite{louis2024interpretable, louis2022statutory,zhong2020jec} are in other languages.

Lastly, few legal IR datasets are paired  with downstream tasks akin to open-domain QA. Thus, few of the available datasets are suitable for end-to-end evaluation of retrieved-augmented LLMs. CLERC \cite{hou2024clerc} includes both a retrieval and retrieved-augmented generation task: given the beginning of a case and retrieved reference paragraphs from cited cases, a model is evaluated on its ability to generate continuing analysis paragraphs of the case. However, as discussed by the authors, automatically measuring factual recall of open-ended text generations is a challenging, unsolved problem \cite{hou2024clerc}. Our datasets are linked to multiple-choice QA tasks. The classification setting makes it easier to automatically evaluate retrieval-augmented LLMs for factual correctness. In law, commons principles or rules are restated many times across the corpus (of case law). In settings where many references passages may be helpful or gold annotations are limited, downstream retrieval-augmented task performance can also be valuable for further contextualizing retrieval performance.

\section{Datasets}
\label{sec:datasets}
Our datasets advance beyond existing open-domain QA datasets and legal IR datasets by offering concrete, substantive legal questions paired with both supporting gold passages and answers.

We highlight the key ways in which our datasets differ from existing datasets. First, our benchmark allows for evaluation of both retrieval and downstream question-answering. In contrast, many prior legal datasets, are either exclusively intended for retrieval evaluation (no associated question-answer pairs) \cite{coliee2024, hou2024clerc}, or intended exclusively for question-answering (no associated document corpora) \cite{guha2024legalbench, zheng2021does, holzenberger2020dataset}. As our paper highlights, datasets that enable evaluation on both tasks allow for a more fine-grained understanding of end-to-end system performance.

Second, our questions and passage labels were hand-annotated by legal experts, bar exam writers and law students for Bar Exam QA and legal researchers for Housing Statute QA. We believe this makes the retrieval task more realistic compared to popular extractive constructions. In these extractive constructions, the query and passage pairs are derived from case citation relationships, where both the query and passage are sections of text extracted from the opinions of the citing case and the cited case. Because of their extractive nature, these datasets simulate the task of retrieving a rule from a cited case using the citing context, but the query is not typically a well-formed question and the passage is not always closely related to the query (due to challenges with localizing the relevant rule within the cited case) \cite{hou2024clerc}. In contrast, our query and passage pairs represent substantive legal questions from multiple areas of law and explanatory rules or passages justifying the answer. To our knowledge, few (if any) English legal retrieval datasets were constructed with hand-annotated passage pairs; existing datasets cover French or Chinese law \cite{louis2024interpretable, louis2022statutory, zhong2020jec}.

Third, our retrieval corpora, particularly for Housing Statute QA, are substantially larger (\textasciitilde 1-2M documents) than the retrieval corpora used in several other legal and general retrieval benchmarks for reasoning intensive retrieval tasks (\textasciitilde 10,000-100,000 documents) \cite{coliee2024, wang2024birco, su2024bright}. Retrieval corpora size matters because retrieval becomes harder to perform as the corpora increases in size and the relative fraction of irrelevant documents increases.

Finally, our datasets focus on specific types of documents and questions, legal rule-application questions over cases (spanning the traditional areas of law tested on the Bar Exam) and housing questions over state statutes, which existing benchmarks do not capture. This is important because the performance of retrievers and LLMs can vary, sometimes significantly, across question and document types.

We recognize our datasets cannot capture the full spectrum of complexities involved in real-world legal tasks. Our datasets are restricted in subject-matter domains and restricted to multiple-choice answer forms to enable automatic evaluation of the downstream task. They may not represent a ``realistic'' approximation of the full natural distribution of legal questions. In particular, the BarExamQA questions are drawn from practice bar exams, where the questions involve stylized, fictional, short fact patterns, which may not be similar to the distribution of real-world fact patterns attorneys encounter. However, these queries advance beyond existing legal retrieval datasets \cite{coliee2024, hou2024clerc} by offering concrete, substantive legal questions and enable comparison to bar exam QA tasks in other popular general reasoning benchmarks \cite{hendrycks2021test, katz2024gpt}. But we acknowledge that as new efforts move towards creating more realistic bar exam questions with multistage factual scenarios, reapplying the techniques described to construct these datasets to those questions would provide even more realistic tasks \cite{mcfarlin2023more}. The Housing Statute QA question are drawn from the LSC Eviction Laws Database, a real-world resource designed to help address tenants' questions related to the legal process of eviction.

We describe the datasets in the benchmark and the process on construction in the following sections. Table \ref{tab:datasets} provides a summary of the datasets. Dataset release and license information is provided in Appendix \ref{sec:dataset_release_licenses}. We show representative examples from each dataset in Appendix \ref{sec:dataset_examples}.

\begin{table*}
    \centering
    \begin{tabular}{lccccc}
         \toprule
         Dataset & \multicolumn{2}{c}{Total Number}  & \multicolumn{2}{c}{Avg. Length} & Examples \\
         \cmidrule(r){2-3} \cmidrule(l){4-5}
         & $\mathcal{\mathbf{Q}}$ & $\mathcal{\mathbf{P}}$ & $\mathcal{\mathbf{Q}}$ & $\mathcal{\mathbf{P}}$ & \\
         \midrule
         \textbf{Ours} \\
         Bar Exam QA & 1,195/1,815 & 856,835 & 172/157 & 131 & Table \ref{tab:ex_barexamqa} \\
         Housing Statute QA & 6,853 & 1,837,403 &  15 & 349 & Table \ref{tab:ex_housingqa} \\
         \midrule
         \textbf{Comparison} \\
         Natural Questions \small{\cite{kwiatkowski2019nq}} & 3,452 & 2,681,468 & 11 & 102 & Table \ref{tab:ex_nq} \\
         HotpotQA \small{\cite{yang2018hotpotqa}} & 7,405 & 5,233,329 & 20 & 63 & Table \ref{tab:ex_hotpotqa} \\
         COLIEE \small{\cite{coliee2024}} & 1,278 & 5,616 & 6,730 & 6,768 & Table \ref{tab:ex_coliee} \\
         CLERC \small{\cite{hou2024clerc}} & 2,851 & 1,842,422 & 415 & 3,303 & Table \ref{tab:ex_clerc} \\
         \bottomrule
    \end{tabular}
    \caption{Summary of datasets. We report number of queries ($\mathcal{\mathbf{Q}}$), number of passages ($\mathcal{\mathbf{P}}$), the average length of queries and passages (calculated with the GPT-2 tokenizer \cite{radford2019language}), and examples. For Bar Exam QA, the query statistics are reported for the Historical MBE/Barbri subsets.}
    \label{tab:datasets}
\end{table*}

\subsection{Bar Exam QA}
\label{sec:bar_exam_qa}
The Bar Exam QA dataset is a dataset of multistate bar exam (MBE) questions. The multistate bar exam is a professional exam that certifies law students to practice law in the U.S. The Bar Exam QA datasets consists of MBE questions from historical bar exams released by the National Conference of Bar Examiners (NCBE) and practice bar exams from Barbri MBE test preparation workbook (2013 Ed.). Each MBE question contains a novel legal scenario, a question about a specific legal issue implicated in the scenario, and four answer choices. The task is to select the correct answer choice.

We transform the dataset into a retrieval task by collecting gold explanation passages for each example. For the Barbri practice bar exams, we extract the explanation passages from the answer key for each question as the gold passage. For the historical bar exams, for which no explanation passages are available, a law student hand-annotated each example with a gold passage that helps or supports the correct answer to the question. The law student's annotation process simulates the legal research process. We provide a detailed description of the annotation process in Appendix \ref{sec:barexamqa_construction}. The authors and research assistants manually validated subsets of the examples and gold passages. Annotations took approximately 6 months for the team to complete.

The retrieval passage pool contains \textasciitilde900K passages. The passage pool consists of the gold passages, U.S. caselaw from 2019-2021 (case decision text split at the paragraph-level), and Cornell Law School Legal Information Institute (LII) Wex legal encyclopedia entries and select primary sources.\footnote{The source documents are segmented at the paragraph-level using this tool: \url{https://github.com/neelguha/legal-segmenter}.}

We release the subset of the dataset containing historical publicly released MBE questions (Historical MBE). We treat the Barbri questions as a private held-out subset and report separate results on this subset (Barbri). Because the historical publicly released MBE questions are contained in the MMLU auxiliary train set for professional law \cite{hendrycks2021test}, these examples may have been used for model training. To our knowledge, the Barbri set has not been previously released in any dataset, and thus, are more likely to be true, unseen examples, for model evaluation. The Barbri set examples also reflect more modern styles of bar exam questions.

\subsection{Housing Statute QA}
\label{sec:housing_statute_qa}
Housing Statute QA is a question answering dataset covering statutory housing law across 50+ U.S. jurisdictions. Each sample in the dataset contains a Yes/No (Y/N) question about housing law in a particular state, the answer to the question, and a small number ($\leq 10$) of ``relevant'' statutes (which contain text support the correct answer). These statutes are mapped to an individual statute in a larger database of state law. In the retrieval setting, the objective is to identify the relevant statutes from the larger database.

Housing Statute QA was created by adapting the Legal Services Corporation (LSC) Eviction Laws Database~\cite{lsc}. The original database was constructed by legally trained researchers and students, who manually answered questions about housing law for different jurisdictions, by explicitly searching housing law in each jurisdiction. Similar to the annotation procedure for Bar Exam QA, the annotation process is modeled off of the legal research process~\cite{lsc}. The database provides questions, answers, and citations to statutes which support the answer.

\begin{figure}
    \centering
    \includegraphics[width=0.8\columnwidth]{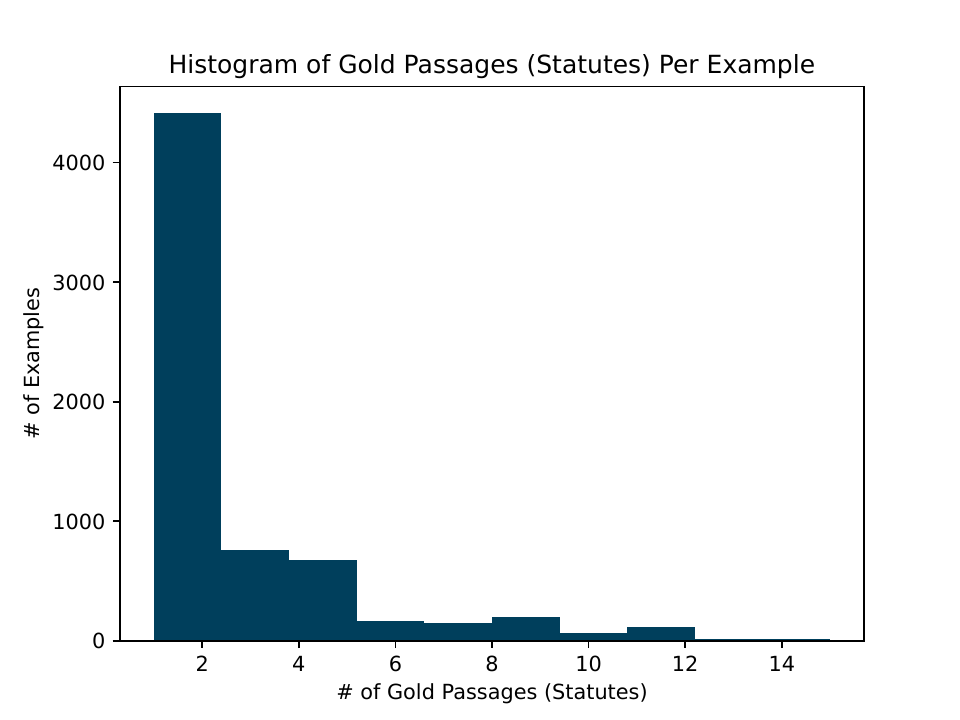}
    \caption{Histogram of number of gold passages (statutes) per example in the Housing Statue QA dataset.}
    \label{fig:statutes_per_ex}
\end{figure}

The original database contains a mixture of free-response, multi-answer multiple-choice, and Y/N questions. Prior work has observed that evaluating LLM responses for non-Y/N responses can be challenging~\cite{guha2024legalbench}. Thus, we restrict Housing Statute QA to only contain Yes/No questions. We do so by first using all the Y/N questions contained in the original LSC database. Next, we convert the multi-answer multiple-choice questions into new Y/N questions. For each answer-choice in multiple-choice answer space, we create a new Y/N question asking if that answer is true. Thus, from a single multiple-choice question with five answer choices, we derive five new Y/N questions. In Appendix \ref{sec:housingqa_construction}, Table \ref{tab:reformatted_housingqa}, we provide an example of an original question from the database and our reformulated Y/N question. In Figure \ref{fig:statutes_per_ex}, we provide a histogram illustrating the distribution of the number of gold passages (statutes) per transformed example in the dataset.

The LSC Database annotates each question with citations to state laws which contain information relevant for answering the question. 
We build a corpus from Justia's available state statutes from the year 2021.\footnote{\url{https://law.justia.com/codes/}} If the 2021 data from the jurisdiction was not available, the most recently published set of statutes was used. We use statute citations on the original questions to identify relevant statutes in this corpora. We note that Justia's coverage of state law is incomplete, and some state statutes are not available via Justia.

Our released version of Housing Statutes QA consists of two splits. The first split (\texttt{rc\_questions})---which we study here---contains 6,853 question-answer pair examples with labeled supporting statutes. This can be used as an evaluation set for the retrieval and downstream question answering tasks. The second split (\texttt{knowledge\_qa}) is larger (9,297 examples), and contains question-answer pairs for which we could not identify a labeled supporting statute. While the lack of a statute annotation prevents these questions being used for retrieval evaluation, we believe they may be of independent interest to researchers. The retrieval passage pool contains \textasciitilde2M passages.

\section{Comparison to Existing Tasks}
\label{sec:task_complexity_analysis}
In typical open-domain question answering tasks, the relevant passages restates or closely restates a significant portion of the question and the answer is, by construction, a substring of the gold passage. Therefore, the relationship between the question, gold passage, and answer for such tasks can often be recovered by comparing lexical similarities of the texts. We find that this is also the case for existing legal IR datasets derived from case citation relationships \cite{coliee2024}. In contrast, our tasks require a greater degree of reasoning to connect the question to the gold passage and answer and the gold passage is more lexically distant from both the question and the answer.

To analyze this property of the tasks, we compare the task complexity of our datasets against two popular general domain IR tasks: Natural Questions (NQ) \cite{kwiatkowski2019nq}, HotpotQA \cite{yang2018hotpotqa}, and two existing legal domain IR tasks: COLIEE (2024) \cite{coliee2024}, CLERC \cite{hou2024clerc}. We use the same metrics to compare task complexity as those used in BIRCO \cite{wang2024birco}, lexical similarity and baseline performance of existing IR methods. We compare the lexical similarities between the (query, gold passage) pair and (gold passage, answer) pair for each dataset example. 

We use TF-IDF cosine similarity as the lexical similarity metric because it is a closely related metric to BM25, a strong lexical baseline ranking function for retrieval \cite{robertson1976relevance}. 
For NQ and HotpotQA, we report metrics over the BEIR benchmark test sets \cite{thakur2021beir}, since these subsets are commonly used to evaluate retrieval performance, the train set of COLIEE (2024)\footnote{We report metrics on Task 1.1, the retrieval task derived from Canadian case law, since these laws were originally written in English, though we find that the metrics on Task 2.1 on the Japanese Civil Code (translated to English) are similar.}, since the test set labels are not released, and the CLERC test set.\footnote{CLERC presents two settings for gold passage selection. The reference passage is either (1) the full case text of the central citation in the query (document) or (2) a set of sampled analysis paragraphs from the case (paragraph). Since it is not clear that an arbitrary section of the full case text supports the specific citing context in a given query, we report scores for the paragraph with maximal lexical similarity.}

\begin{figure}[]
    \begin{subfigure}[t]{\columnwidth}
        \centering
        \includegraphics[scale=0.5]{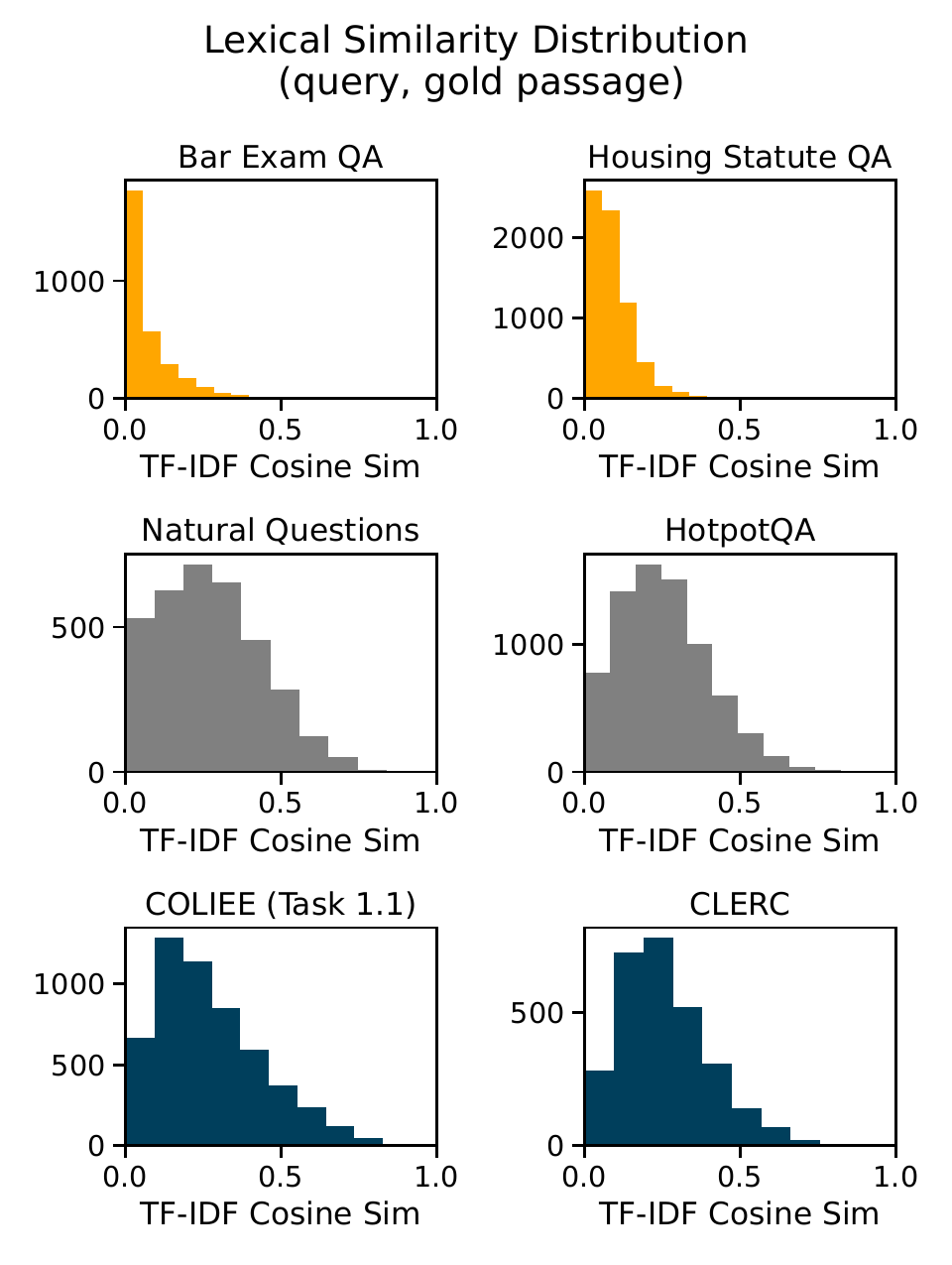}
        \label{fig:qgp_lex_sims_histograms}
    \end{subfigure} \\
    \begin{subfigure}[t]{\columnwidth}
        \centering
        \includegraphics[scale=0.5]{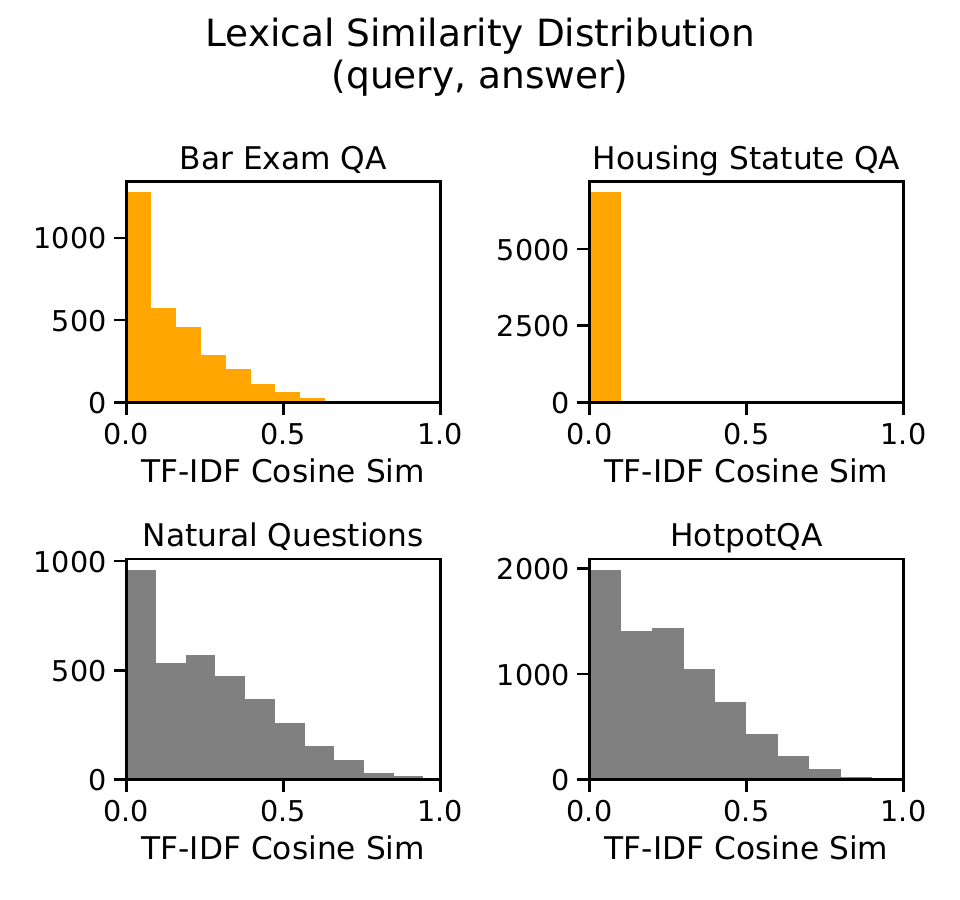}
        \label{fig:gpa_lex_sims_histograms}
    \end{subfigure}
    \caption{Histograms of the example lexical similarity of (query, gold passage) and (gold passage, answer) over the following datasets: Bar Exam QA and Housing Statute QA (row 1), NQ and HotpotQA (row 2), COLIEE and CLERC (row 3). In Appendix \ref{sec:lex_sim_stat_test_results}, we report results for the Kolmogorov-Smirnov test for distributional equivalence between the task similarity distributions, which provide evidence of statistical significant differences between our datasets and other representative datasets at $\alpha = 0.05$.} 
    \label{fig:lex_sims_histograms}
\end{figure}

Figure \ref{fig:lex_sims_histograms} shows that the lexical similarities between query and gold passage for NQ, HotpotQA, COLIEE, and CLERC are distributed normally around a mean of 0.25 - 0.27 (Table \ref{tab:lex_sim}), while those distributions for Bar Exam QA and Housing Statute QA are heavily skewed towards similarities < 0.10, with mean similarities of 0.07 and 0.08 (Table \ref{tab:lex_sim}). Lexical similarities are also lower for the gold passages and answers in our tasks, since additional inference is typically needed to conclude the correct legal outcome in the answer from the gold passage rules and the facts in the query.\footnote{We note that for Housing Statute QA, question types with categorical answers are transformed to Yes/No answers to standardize downstream evaluation, so the lexical similarity of the original answer is likely higher for this subset of questions.}


\begin{table}[t]
    \centering
    \begin{tabular}{lcc}
        \toprule
        Dataset & \thead{$LS_{retrieval}$} & \thead{$LS_{QA}$} \\
        \midrule
        Bar Exam QA & $0.07 \pm 0.00$ & $0.14 \pm 0.01 $ \\
        Housing Statute QA & $0.09 \pm 0.00$ & $0.00 \pm 0.00$ \\
        \midrule
        Natural Questions & $0.27 \pm 0.01$ & $0.25 \pm 0.01$ \\
        HotPotQA & $0.26 \pm 0.00$ & $0.24 \pm 0.00$ \\
        COLIEE & $0.27 \pm 0.00$ & - \\
        CLERC & $0.26 \pm 0.01$ & - \\
        \bottomrule
    \end{tabular}
    \caption{Lexical similarity task score for the retrieval task and downstream QA task for each dataset. $LS_{retrieval}$ is the mean lexical similarity score between (query, gold passage) and $LS_{QA}$ is the mean lexical similarity score between (gold passage, answer) over dataset examples, reported with 95\% confidence intervals reported. In Appendix \ref{sec:lex_sim_stat_test_results}, we report results for the t-test for the difference of means, which provide evidence of statistical significant differences between our datasets and other representative datasets at $\alpha = 0.05$.}
    \label{tab:lex_sim}
\end{table}

\begin{table}[]
    \centering
    \begin{tabular}{lrr}
         \toprule
         Dataset & BM25 & E5-large-v2 \\
         \midrule
         Bar Exam QA & 5.03 & 7.00 \\
         Housing Statute QA (lower) & 18.3 & 24.4 \\
         Housing Statute QA (upper) & 40.8 & 50.6 \\
         \midrule
         Natural Questions & 40.4 & 68.7 \\
         HotpotQA & 32.7 & 56.2 \\
         COLIEE & 38.1 & 32.7 \\
         CLERC & 11.8 & 6.80 \\
         \bottomrule
    \end{tabular}
    \caption{Baseline retrieval performance (Recall@10) of BM25 (lexical) and E5-large-v2 (dense) retrieval methods on Bar Exam QA (aggregate), Housing Statute QA, NQ, HotpotQA, COLIEE, and CLERC. We report the recall lower/upper bound for Housing Statute QA, see Section \ref{sec:experimental_setup} for details.}
    \label{tab:baseline_metrics_comparison}
\end{table}



\section{Evaluation}
\label{sec:baselines}
\subsection{Baseline Retrievers}
We evaluate a number of baselines, including BM25 \cite{robertson1976relevance}, which has been shown to be a robust lexical retrieval baseline \cite{lin2019neural, thakur2021beir, rosa2021yes}, and the E5 family of retrieval models: E5-small-v2, E5-base-v2, E5-large-v2, E5-mistral-7b-instruct, a series of dense embedding models available in a range of sizes that have been shown to perform well on a broad suite of tasks and complex retrieval tasks in particular \cite{wang2022text, wang2023improving, wang2024birco}. The E5 models are trained from MiniLM \cite{wang2020minilm}, BERT base and large (uncased) \cite{devlin-etal-2019-bert}, and Mistral 7B Instruct \cite{jiang2023mistral} models.

\subsection{Experimental Setup}
\label{sec:experimental_setup}
For Bar Exam QA, we evaluate retrieval performance on the full passage corpus. In the body of the paper, we report results for the aggregate dataset. We report disaggregated results for the Historical MBE and Barbri subsets in Appendix \ref{sec:comp_retrieval_results}.

For Housing Statute QA, the dataset includes information about the jurisdiction (U.S. state or territory) of each query and passage, so for each query, we retrieve from a candidate passage pool of the statute passages for the given jurisdiction. The candidate passage pools for each jurisdiction range in size from 10,676 to 155,974 passages. Due to the transformation of the original database questions to Y/N questions described in Section \ref{sec:housing_statute_qa}, not all of the gold passages for the original question may be relevant to each Y/N question. However, we show in Figure \ref{fig:statutes_per_ex} that the vast majority of the transformed examples have only 1-2 gold passage labels. We report recall as the retrieval of at least one gold passage for a given query (upper bound) in Results (Section \ref{sec:results}). We include full retrieval results computing recall as the retrieval of all the gold passages for a given query (lower bound) in Appendix \ref{sec:comp_retrieval_results}.

We also evaluate the comparison tasks on the same baselines. 


\subsection{Query Expansion}
\label{sec:expansion}
As discussed in \citet{wang2024birco, jia2023mill}, simple retrieval methods can fail to capture the correct search intent or task objective when the retrieval request itself requires reasoning---often the case in legal retrieval-augmented QA. However, recent work on query expansion~\citep{wang2024birco,lei2024corpus,jagerman2023query} may provide some path forward.
To that end, we test several retrieval methods for query expansion in addition to baseline retrieval methods. We use GPT-3.5 (gpt-3.5-turbo-0613) as the generative model for our query expansion experiments.\footnote{We set maximum length = 1024 and temperature = 1 and use the default hyperparameters otherwise (top p = 1, frequency penalty = 0, presence penalty
= 0, best of = 1).}

\paragraph{Paraphrasing.} 
We evaluate query expansion using a prompt to the generative model to paraphrase the query. As illustrated in Table \ref{tab:ex_barexamqa}, the queries for Bar Exam QA are often quite long. We use this prompting method to study whether simplifying the language in the query helps the retriever. The queries in Housing Statute QA are short, so we do not test this query expansion method for that dataset.

\paragraph{Chain-of-Thought (CoT).} \citet{jagerman2023query} use CoT prompting to expand user queries and find performance increases on BEIR \cite{thakur2021beir}, we compare against their method. We also evaluate the comparison tasks on this reasoning query expansion method.

\paragraph{Structured Legal Reasoning.} We build on these prior approaches and also test a modified query expansion method tailored to the legal setting. Our query expansion method prompts a generative model to perform structured reasoning about the relevant higher-order knowledge hierarchy of the legal task (e.g., the higher-level rules implicated by the query facts) and expands the query with the generated reasoning rollout. The closest prompt-based query expansion approach to this is that of \citet{jagerman2023query}, however, we note that ours encodes the legal reasoning process by explicitly prompting the generative model to perform legal issue spotting and brainstorm potential legal rules that address the issue. 
In some ways, this is also related to the prompt-based task-specific re-ranking method by \citet{wang2024birco}, since it adds task-specific prompting and domain knowledge to the retrieval mechanism, but their approach focuses on re-ranking rather than query expansion.

The exact prompts for the query expansion methods on Bar Exam QA and Housing Statute QA are available in Appendix \ref{sec:query_expansion_prompts}. In Appendix \ref{sec:structured_reasoning_prompt_encodes}, we show an example of the generated query expansion for the same question with the different prompting methods, to illustrate how the structured reasoning rollouts encode the implicit steps required for the legal retrieval tasks by capturing the latent issues and enumerating potential rules addressing the issues that match the language of the statements of law (or primary sources of law) in the passage corpora.



\begin{figure*}[]
    \centering
    \includegraphics[width=\linewidth]{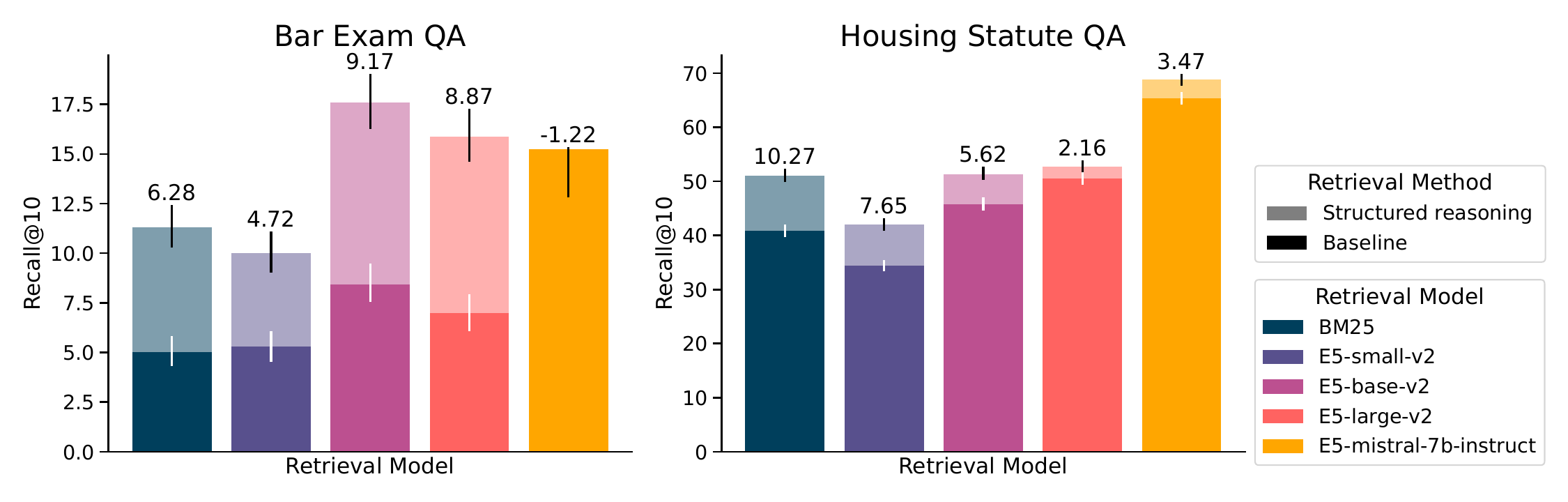}
    \caption{Recall of baseline and structured reasoning rollout query expansion retrieval for lexical (BM25) and dense models (E5 family), evaluated on our legal retrieval benchmark tasks. The gain in Recall@10 with the structured reasoning rollout method is labeled above each bar. 95\% confidence intervals are estimated with a percentile bootstrap ($n = 1000$). For Housing Statute QA, recall is reported as retrieval of at least one gold passage (upper bound).}
    \label{fig:gain_recall}
\end{figure*}

\subsection{Retrieval-Augmented Question Answering}
We evaluate downstream QA performance for Llama 3 8B Instruct\footnote{\url{https://ai.meta.com/blog/meta-llama-3/}} and GPT-4o-mini (gpt-4o-mini-2024-07-18) on Bar Exam QA and Housing Statute QA. We evaluate baseline performance with no passage, performance with retrieved passages using each baseline retriever (with and without each query expansion method), performance with the generative reasoning rollout from the structured legal reasoning query expansion method as a pseudo-passage, and performance with the annotated gold passages.

For Llama 3 8B Instruct, we predict the answer by taking the maximum likelihood prediction over the answer choice letters (e.g., A, B, C, D for Bar Exam QA, Y or N for Housing Statute QA).\footnote{As in the implementation here: \url{https://github.com/artidoro/qlora/blob/main/qlora.py}} For GPT-4o-mini, we predict the answer with an open-ended generative setup, using the default hyperparameters set by the API.

\section{Results}
\label{sec:results}
\textbf{Bar Exam QA and Housing Statute QA are challenging for baseline retrievers, especially lexical retrievers like BM25 and smaller models.} As noted by \citet{wang2024birco}, it is difficult to directly compare retrieval metrics across datasets due to differences in the relevance scale and number of relevant passages per query across tasks, but we report baseline retrieval performance on our tasks and the comparison tasks in Table \ref{tab:baseline_metrics_comparison} and Appendix \ref{sec:comp_retrieval_results}.
In particular, BM25, a strong lexical baseline \cite{robertson1976relevance}, and E5-large-v2, a dense embedding retrieval model that BIRCO evaluation focuses on due to its strong performance on complex tasks \cite{wang2022text, wang2024birco}, achieve significantly lower recall on Bar Exam QA than other tasks. In comparison, BM25 performs well for NQ and HotpotQA and can be a surprisingly strong baseline for some other legal IR tasks like COLIEE, due to high lexical similarity between queries and gold passages (Table \ref{tab:lex_sim}).\footnote{For CLERC, we evaluate the document-level setting. In general, in a legal citation, the full cited case document is not necessarily relevant to the citing context. Citing contexts typically refer to a specific section of the cited case as support. Baseline models would likely yield higher performance on more granular section-level annotations.}

\textbf{Query expansion, particularly our structured query expansion, helps lexical retrievers and small models recover some performance on these legal tasks.} Figure \ref{fig:gain_recall} shows baseline retrieval performance without query expansion and retrieval performance with structured legal reasoning query expansion. For our legal tasks, characterized by reasoning-focused retrieval task objectives and low (query, relevant passage) lexical similarity, our method of generative query expansion achieves statistically significantly improvement on baseline retrieval performance across all models evaluated, with the largest gains in recall for the more lexically-focused retrieval models. On Bar Exam QA, the gain in Recall@10 for structured reasoning query expansion over baseline retrieval is $6.28 \pm 0.99$ for BM25 and $8.86 \pm 1.16$ for E5-large-v2. On Housing Statute QA, the gain in Recall@10 for structured reasoning query expansion over baseline retrieval is $10.27 \pm 1.08$ for BM25 and $2.16 \pm 1.15$ for E5-large-v2. The gains in performance are statistically significant for BM25 and the three smaller E5 models. We do not observe gains for E5-mistral-7b-instruct, but this model is also already highly performant at baseline. We believe this is likely because E5-mistral-7b-instruct is fine-tuned on instruction-following data for retrieval tasks \cite{wang2023improving}, so it may have greater learned retrieval task objective awareness than BM25 or smaller dense embedding models.


Structured reasoning rollouts outperform other prompting techniques for generative query expansion that increase the verbosity of the query. 
Paraphrasing does not improve and can hurt retrieval performance slightly compared to baseline on Bar Exam QA, 
suggesting that summarizing long queries and introducing synonymous language may not be sufficient for improving retrieval performance on legal retrieval tasks that necessitate additional reasoning about the query. 

We find that CoT reasoning query expansion improves retrieval performance over baseline on all models for Bar Exam QA. 
For Bar Exam QA, where CoT is more effective, the difference between CoT and structured reasoning is smaller, but still significant for most models at Recall@10 ($3.09 \pm 1.15$ for E5-large-v2). For Housing Statute QA, where CoT is less effective, the difference beween CoT and structured reasoning in larger and significant for all models at Recall@10 ($8.97 \pm 1.01$ for E5-large-v2).

For datasets with higher lexical similarity between questions and gold passages, reasoning rollouts for query expansion are less helpful. We hypothesize that this is because less reasoning is required to complete the retrieval task, so existing retrieval models at baseline already achieve strong performance. For Bar Exam QA, the gain in Recall@10 for CoT over baseline retrieval for E5-base-v2 is $7.37 \pm 1.15$, while on NQ and HotpotQA, the difference is $1.59 \pm 1.45$ and $-2.79 \pm 1.12$ respectively. This suggests the expected gain of generative reasoning rollouts for query expansion on retrieval tasks may be greater for low lexical similarity tasks compared to high lexical similarity tasks. For some high lexical similarity retrieval benchmarks, such as NQ, current state of the art retrieval models at the \textasciitilde100-300M parameter size approach performance saturation on the benchmark. Retrieval tasks with lower lexical similarity that necessitate greater reasoning, such as ours, are more challenging for these models, and we observe the greatest gains from appending more reasoning tokens through query expansion in these cases. Additionally, performance may depend on the quality of the generative reasoning rollout. Though we use one generative model, GPT-3.5, in our experiments, we expect that improvement will be correlated with the quality of the generative model used for reasoning rollout.

\begin{figure}[h]
    \centering
    \includegraphics[width=\columnwidth]{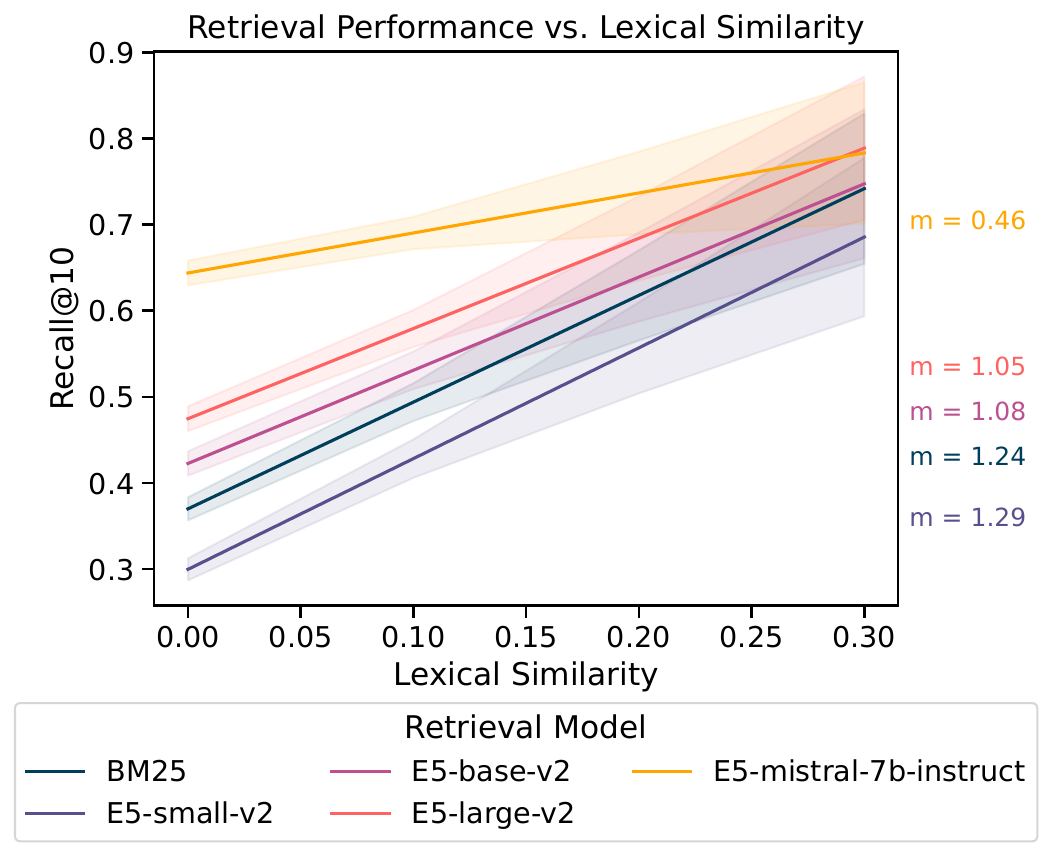}
    \caption{Baseline retrieval performance vs. lexical similarity (query, gold passage) line of best fit over Housing Statute QA task. Recall@10 is averaged over examples bucketed by intervals of lexical similarity scores (bucket sizes of 0.1). 95\% confidence intervals are estimated over each bucket. $m$ is the slope of the line of best fit.} 
    \label{fig:recall_ls}
\end{figure}

\textbf{Hard retrieval task examples help distinguish more capable retrieval models.}
Figure \ref{fig:recall_ls} shows the relationship between baseline retrieval performance and example query and gold passage lexical similarity for the Housing Statute QA dataset, across the five retrieval models. We observe that for examples with high query and gold passage similarity, models perform similarly well. The E5-mistral-7b-instruct model outperforms the other models most significantly on the set of hard examples with low query and gold passage similarity. The relationship illustrates the importance of more complex retrieval tasks for benchmarking retrieval model performance.

\textbf{Improvement in retrieval performance (Recall@10) doesn’t always translate to significant  downstream improvements in QA.} One challenge is that the retriever must be able to reason about which passages to retrieve, but the downstream LLM must also be able to reason about the retrieved passages. We report full results for downstream evaluation with no passage, retrieved passages, the generative reasoning rollout from the structured reasoning query expansion method as a pseudo-passage, and the gold passage on Llama 3 8B Instruct and GPT-4o-mini in Appendix \ref{sec:downstream_qa_results}. We find that improvements are upper-bounded by how well models can make use of the gold passage, with only a 20\% gain for Llama 3 8B Instruct. 
However, because the maximum improvement from finding the optimal passage is 20\%, even a 10\% gain on retrieval can only lead to a 2\% improvement on downstream task performance in theory, consistent with the trends that we see (Appendix \ref{sec:downstream_qa_results}). 

On GPT-4o-mini (gpt-4o-mini-2024-07-18), we find that the model still struggles to reason over and apply both the retrieved and gold passages to the more challenging Bar Exam QA questions. Using the generative reasoning rollout as a pseudo-passage seems to confuse the model, resulting in a reduction in accuracy, likely from distractors in the generated information. On the other hand, GPT-4o-mini achieve significant improvement in accuracy with retrieved passages over no passage on the downstream task for Housing Statute QA (23.53 percentage point gain). On this case, the generative reasoning rollout as a pseudo-passage also helps improve downstream QA performance, but not as much as using the retrieved passage (68.51\% vs. 71.71\% accuracy). This suggests that a more capable downstream reasoning model can be helpful for evaluating the utility of retrieved passages to difficult reasoning QA tasks and distinguishing between the quality of various passage sources.

To improve legal retrieval augmented LLMs, future work should focus on improving the reasoning abilities of retrievers as well as the ability of downstream models to reason about retrieved passages.

\section{Conclusion}
In this work, we introduce two new benchmark datasets for evaluating retrieval-augmented question answering in the legal setting: Bar Exam QA and Housing Statute QA. This benchmark provides \textasciitilde10K paired, query, gold passage, answer examples, with high-quality, human-annotated gold passages. These datasets contain substantive legal questions as queries and supporting law as passages, simulating reasoning-intensive real-world legal retrieval tasks. 

We note that Bar Exam QA and Housing Statute QA do not represent the full distribution of legal questions legal practitioners are likely to encounter in practice, since they cover only the areas tested on the Bar Exam and statutory housing law. However, they provide a closer setting to real-world legal tasks than many other legal retrieval datasets. Unlike legal retrieval datasets with extractive constructions, our query, gold passage pairs were hand-annotated by law students, who were instructed to use legal research tools to find supporting law for a given legal question justifying the answer.

Our benchmarks serve to help researchers, practioners, and policymakers better understand the suitability of retrieval approaches for different legal retrieval tasks over time, as models' performance on general-domain retrieval benchmarks don't necessarily appear to generalize well to law. In our evaluations, we show that the tasks are challenging for lexically-focused retrievers, but generative query expansion techniques that roll out reasoning can help improve retrieval performance. These findings suggest that retrievers must themselves be reasoners too. And that certain legal tasks may be particularly well suited to exposing limitations of current retrieval models on reasoning-intensive retrieval tasks. This conclusion comports with discussions among legal scholars that determining what law is relevant to addressing a legal question is itself a nontrivial problem and is a separate reasoning skill from the reasoning skills required to apply relevant law to novel scenarios \cite{curcio2018build}. We hope that our datasets and evaluations can serve as a resource for future work on reasoning-focused retrieval-augmented LLM tasks.

\begin{acks}
We thank Maura Carey for annotating gold passages for the Bar Exam QA dataset. We thank Yvonne Hong for research assistance in processing the Bar Exam QA dataset. We thank Isaac Cui, Olivia Martin, and Catherina Xu for piloting early iterations of the retrieval method. We are grateful to Varun Magesh, Faiz Surani, Suvir Mirchandani, Isabel Gallegos, Jihyeon Je, Chenglei Si, and Aryaman Arora for helpful discussion.

LZ is supported by the Stanford Interdisciplinary Graduate Fellowship (SIGF). NG is supported by the Stanford Interdisciplinary Graduate Fellowship (SIGF) and the HAI Graduate Fellowship.

This work is dedicated to Andrea Vallebueno, in loving memory. Andrea was a dear friend and labmate. She had a brilliant, warm spirit with a special gift for research and teaching others. Her light overflowed on to each person in her life and inspired so so many, close and far. We remember and hope to carry on the legacy of her life, the dignity and respect with which she treated every person she encountered, her welcoming and inclusive nature, and her passion for her research on computational methods for addressing socially impactful problems.
\end{acks}

\bibliographystyle{ACM-Reference-Format}
\bibliography{main}

\appendix

\section{Bar Exam QA Dataset Construction}
\label{sec:barexamqa_construction}
The gold passage annotation process for Bar Exam QA is modeled off of the legal research process. For each example, the law student has access to the general area of law of the question, the question, the answer choices, and the correct answer. This annotation effort took roughly 9 months to complete.

\begin{enumerate}
    \item First, the law students identify the rule of law relevant to the example. They consider the Barbri answer key bank (from the set of examples with explanation passages), bar exam study guides and practice guides (e.g., \textit{The Ultimate Guide to the MBE}), and other secondary sources (e.g., American Law Reports). They identify general rules of law that could help answer the question and a list of general and specific search terms and legal concepts based on keywords from these secondary sources.
    \item  Next, the law students compose a Westlaw Terms and Connector search query for relevant cases stating the rule of law. These search queries are written by hand by the law students, without AI assistance. In Table \ref{tab:exs_terms_and_connector_queries}, we provide examples of Terms and Connector search queries constructed by law student annotators in the search process. From the search results, the law students read the descriptions of cases for one that appears on point and review the case headnotes to identify a case with a statement of the rule of law that mirrors the rule of law identified.
    \item Lastly, the law students find a succinct, generalizable statement of the rule of law in the identified case text and annotates the example with this text as the gold passage label.
\end{enumerate}

\begin{table}[h]
    \centering
    \begin{tabular}{c}
        \toprule
        "due process" /p "termination" \\
        "unreasonable burden on interstate commerce" \\
        controvers! /p "declaratory judgment" \\
        “Presidential pardon power" \\
        “Due process” +5 “balancing” \\
        \bottomrule
    \end{tabular}
    \caption{Examples of Westlaw Terms and Connector search queries}
    \label{tab:exs_terms_and_connector_queries}
\end{table}

\section{Housing Statute QA Dataset Construction}
\label{sec:housingqa_construction}
Table \ref{tab:housing_qa_sample_questions} shows a sample of original questions from the LSC Eviction Database \cite{lsc}. Table \ref{tab:reformatted_housingqa} shows examples of reformatted original questions to Y/N questions.

\begin{table*}[p]
    \small
    \centering
        \begin{tabularx}{\textwidth}{X} 
        \toprule
            Can a landlord evict a tenant for endangering property? This includes where the law refers to situations that could result in damage to the property; or a pet capable of causing damage to persons or property. \\ \midrule
            Secondary methods of service are defined as those methods that may be used if the primary method is unsuccessful. Are certified mail and regular mail a permitted  secondary methods of service for an eviction action? \\ \midrule 
            Are court records for eviction cases by default publicly available? \\ \midrule
            Does the law require landlords to provide information on how to cure when giving tenants notice to vacate the property? \\ \midrule 
            Can a landlord evict a tenant for committing or failing to dispose of waste?
\\ \bottomrule
        \end{tabularx}
    \caption{Sample questions for Housing Statutes QA}
    \label{tab:housing_qa_sample_questions}
\end{table*}

\begin{table*}[p]
    \small
    \centering
        \begin{tabularx}{\textwidth}{XXX} 
        \toprule
            \textbf{Original question} & \textbf{Original answer choices} & \textbf{Reformatted Y/N questions}\\ \midrule
            What type(s) of landlord(s) does state/territory eviction law explicitly regulate? & Residential landlords generally \newline\newline Mobile/manufactured home landlords \newline\newline Corporate landlords \newline\newline Floating home landlords \newline\newline Landlords with minimal rental properties & Are residential landlords explicitly regulated by eviction law?\newline\newline Are mobile/manufactured home landlords explicitly regulated by eviction law?\newline\newline Are corporate landlords explicitly regulated by eviction law? \newline\newline Are floating home landlords explicitly regulated by eviction law?\newline\newline Are landlords with minimal rental properties explicitly regulated by eviction law?
\\ \bottomrule
        \end{tabularx}
    \caption{Example of reformatted original questions to Y/N questions}
    \label{tab:reformatted_housingqa}
\end{table*}

\section{Dataset Release and Licenses}
\label{sec:dataset_release_licenses}
Our datasets are publicly available on HuggingFace at the following links.
\begin{itemize}
    \item Datasets: \url{https://huggingface.co/collections/reglab/a-reasoning-focused-legal-retrieval-benchmark-67a00c363f7e0d14619e95c5}
    \begin{itemize}
        \item Bar Exam QA: \url{https://huggingface.co/datasets/reglab/barexam_qa}
        \item Housing Statute QA: \url{https://huggingface.co/datasets/reglab/housing_qa}
    \end{itemize}
\end{itemize}

The passage pool comes from 3 sources with permissive licenses: Cornell LII (CC BY-NC-SA 2.5), Case Law (Public Domain), Justia (Public Domain). Our compilation of these sources follows these licenses.

For Bar Exam QA, for the historical MBE subset, we release the queries and gold passages, annotated by our research team, which we believe are a transformative fair use of the queries, and release them under a CC-BY-NC-SA license. We also believe that release of the historical bar exam multiple choice options and answers in full would be a transformative fair use, since it is for public interest educational purposes, unlikely to affect markets for exams (since they are older and no longer for sale), and much of the data can already be found in other fair use compilations like MMLU professional law auxiliary training set \cite{hendrycks2021test}, Common Crawl, and others. However, in the interest of responsible practice, we release the multiple choice options and answers only to researchers through a gated release mechanism with a restrictive license to the compilation. Some of these can nonetheless already be found in the MMLU auxiliary training set \cite{hendrycks2021test} released under an MIT license, or can be acquired from publicly available sources on the web, available in CommonCrawl and Archive.org. For the Barbri subset, we do not release the dataset due to copyright concerns and treat the subset as a private, held-out test set, which we report separate evaluation results on.

For Housing Statute QA, we release under CC-BY-SA. LSC allows download and redistribution.

\section{Dataset Examples}
\label{sec:dataset_examples}
Tables \ref{tab:ex_barexamqa}, \ref{tab:ex_housingqa}, \ref{tab:ex_nq}, \ref{tab:ex_hotpotqa}, \ref{tab:ex_coliee}, and \ref{tab:ex_clerc} show representative examples from Bar Exam QA, Housing Statute QA, Natural Questions, HotpotQA, COLIEE (Task 1.1), and CLERC.

\begin{table*}[p]
    \centering
    \begin{tabularx}{\linewidth}{lX}
         \toprule
         Query &  \small Under an aid-to-education statute passed by the state legislature a few years ago, many private schools receive state benefits. One private school receives: (i) free textbooks from the state, (ii) an exemption from state taxes, and (iii) 20\% of its operating budget in the form of state grants. The remaining 80\% of the school's-budget is covered by tuition fees and by donations from alumni and others. The school is licensed by the state, but the state has no requirement for certification and licensure of teachers in private schools. 

         A teacher was hired to teach history at the school. The teacher was given the standard three-year contract given to teachers in their first stint at the school. In the fall term of his second year, the teacher gave a lecture to his students criticizing the school's use of school uniforms and encouraging the students to organize a protest against the uniform policy. After the speech, the teacher was called to the administrative office by the headmaster and fired on the spot, despite the teacher's protests that he had almost two years left on his contract. The teacher requested a hearing and was told to leave the premises of the school immediately. 

         If the teacher files suit in federal district court alleging that his constitutional rights have been violated, the teacher will:\\
         Gold Passage & \small The Fourteenth Amendment Due Process Clause, which makes many of the provisions of the Bill of Rights applicable to the states, does not apply to purely private conduct that interferes with these rights. Thus, unless the private individual (i) was performing exclusively public functions, or (ii) took actions with significant state involvement, the individual's action is not unconstitutional. In this case, the school is a private institution performing a function-education-that has never been considered to be an exclusively public function. [See Pierce v. Society of Sisters (1925)] Furthermore, its licensing by the state and receipt of state funds do not constitute significant state involvement with regard to its personnel matters. [See Rendell-Baker v. Kohn (1982)] \\
         Answer & \small Fail, because assistance and involvement by the state did not result in the private school's action being conduct by the state. \\
         \bottomrule
    \end{tabularx}
    \caption{Example from Bar Exam QA}
    \label{tab:ex_barexamqa}
\end{table*}

\begin{table*}[p]
    \centering
    \begin{tabularx}{\linewidth}{lX}
         \toprule
         Jurisdiction & \small Alabama \\
         Query & \small Does the law specify rebuttals available to tenants subject to eviction proceedings? \\
         Gold Passage(s) & \small (b) If a landlord acts in violation of subsection (a), the tenant is entitled to the remedies provided in Section 35-9A-407 and has a defense in any retaliatory action against the tenant for possession. \\
         & \small (a) In an action for possession or in an action for rent when the tenant is in possession, the tenant may counterclaim for any amount the tenant may recover under the rental agreement or this chapter. It is in the court's discretion whether the tenant is to remain in possession. The tenant shall pay into court rent accrued and thereafter accruing as it comes due. The court shall determine the amount due to each party. The party to whom a net amount is owed shall be paid first from the money paid into court, and the balance by the other party. If no rent remains due after application of this section, judgment shall be entered for the tenant in the action for possession. If the defense or counterclaim by the tenant is without merit and is not raised in good faith, the landlord may recover reasonable attorney's fees. \\
         & \small Acceptance of rent with knowledge of a default by the tenant or acceptance of performance by the tenant that varies from the terms of the rental agreement constitutes a waiver of the landlord's right to terminate the rental agreement for that breach, unless otherwise agreed after the breach has occurred. \\
         & \small (b) If contrary to the rental agreement or Section 35-9A-204, after receiving notice of the breach from the tenant, the landlord willfully or negligently fails to promptly make available heat, running water, hot water, electric, gas, or other essential service, the tenant may:(1) send a written notice specifying the date of termination not less than 14 days after receipt of notice and upon vacation of the premises, the rental agreement shall be rightfully terminated without further obligation or penalty. If the rental agreement is terminated pursuant to this section, the landlord shall return all security recoverable by the tenant under Section 35-9A-201 and all unearned prepaid rent; or(2) recover damages based upon the diminution in the fair rental value of the dwelling unit. \\
         & \small (a) Except as provided in this chapter, if there is a material noncompliance by the landlord with the rental agreement or a noncompliance with Section 35-9A-204 materially affecting health and safety, the tenant may deliver a written notice to the landlord specifying the acts and omissions constituting the breach and that the rental agreement will terminate upon a date not less than 14 days after receipt of the notice if the breach is not remedied within that period, and the rental agreement shall terminate as provided in the notice subject to the following: \\
         Answer & \small Yes \\
         \bottomrule
    \end{tabularx}
    \caption{Example from Housing Statute QA}
    \label{tab:ex_housingqa}
\end{table*}

\begin{table*}[p]
    \centering
    \begin{tabularx}{\linewidth}{lX}
         \toprule
         Query &  \small Where is the bowling hall of fame located? \\
         Gold Passage & \small The World Bowling Writers ( WBW ) International Bowling Hall of Fame was established in 1993 and is located in the International Bowling Museum and Hall of Fame , on the International Bowling Campus in Arlington , Texas. \\
         Answer & Arlington , Texas \\
         \bottomrule
    \end{tabularx}
    \caption{Example from Natural Questions \cite{kwiatkowski2019nq}}
    \label{tab:ex_nq}
\end{table*}

\begin{table*}[p]
    \centering
    \begin{tabularx}{\linewidth}{lX}
         \toprule
         Query & \small Peter Marc Jacobson is best known as the co-creator of the popular sitcom "The Nanny", which he created and wrote with his then wife an actress born in which year? \\
         Gold Passage (s) & \small Peter Marc Jacobson (born October 27, 1957) is an American television writer, director and producer, and actor. He is best known as the co-creator of the popular sitcom "The Nanny", which he created and wrote with his then wife actress Fran Drescher, who was the star of the series. He was often credited as Peter Marc in his early acting roles.\\
         & \small Francine Joy "Fran" Drescher (born September 30, 1957) is an American actress and activist. She is best known for her role as Fran Fine in the hit TV series "The Nanny" (1993–99), and for her nasal voice and thick New York accent. \\
         Answer & 1957 \\
         \bottomrule
    \end{tabularx}
    \caption{Example from HotpotQA \cite{yang2018hotpotqa}}
    \label{tab:ex_hotpotqa}
\end{table*}

\begin{table*}[p]
    \centering
    \begin{tabularx}{\linewidth}{lX}
        \toprule
        Query & \small Summary: 
 
        The applicant, a citizen of Afghanistan, claimed refugee protection. In 2010, he was accepted for resettlement to Canada as a member of a humanitarian-protected person abroad class (country of asylum). In 2013, the Minister of Public Safety and Emergency Preparedness and the Minister of Citizenship and Immigration applied for an order that the applicant's refugee status cease on the basis that he had reavailed himself of the protection of his country of nationality (Immigration and Refugee Protection Act, s. 108(1)(a)). The Refugee Protection Division allowed the application. The applicant applied for judicial review. 
 
        The Federal Court dismissed the application and certified the following question: "In a cessation application pursuant to paragraph 108(1)(a) of IRPA, do the same or substantially the same legal considerations, precedents, and analysis apply to persons found to be Convention refugees as to persons found to be in need of protection as members of the Country of asylum class?" ...\\
         Gold Passage(s) & \small  [30] 
        Ample case law from the Immigration Appeal Division in the 1980s was directly concerned with a child's intent in a context where a child's parents had abandoned permanent residence. Although these decisions are not binding on the Court ( 
        $\text{<FRAGMENT\_SUPPRESSED>}$  (QL/Lexis) at paragraph 14), they emphasize the importance of considering the intention of minors when they reach the age to form it, on the basis that they could not form it upon the departure of their parents because of their young age. 
 
        [31] 
        The male applicant was three years old at the time of the initial departure to Mexico and was therefore not able not form an intention to reavail himself of the protection of Mexico. This could have been different at eleven years of age, his age at the time of the hearing. At that point, there should have been further analysis in order to find that an 11-year-old child cannot form an intention that differs from that of his parents. 
         
        [32] 
        However, nothing in the evidence or in the submissions made by the parties makes it possible to determine whether the intention of the child could have been different from that of his mother. 

        IX.  
        Conclusion 
        [33] 
        In the circumstances of this case and in light of the foregoing, the Court cannot intervene because the decision does not go beyond the range of reasonableness...\\
        & \small  [22] 
        However, the Court finds that the Board erred in its consideration of the applicant's explanation relating to his business activities in Thailand. As outlined in  
        $\text{<FRAGMENT\_SUPPRESSED>}$ , a review on the standard of reasonableness is concerned with the "existence of justification, transparency and intelligibility" in the decision. With respect, the Court finds a justification lacking in the present case. It is unclear to the Court why the Board believed that the applicant's explanation with respect to why he obtained a Congolese passport was insufficient. This conclusion may have been open to the Board to make; however, the Court finds it unreasonable that the Board failed to indicate why this explanation was insufficient. If the Board did not believe the applicant's explanation and found him not to be credible then it should have said so. If it had another reason for not finding the explanation sufficient, it should have stated so as well, especially with the type of explanations provided here by the applicant to rebut his presumed intention "to avail himself of the protection of the country of his nationality". 
       [23] 
       True the burden was on the applicant to rebut this presumption, and he tried. But here his explanations as a whole were not discarded by the Board because they were not credible; on the contrary the decision seems to imply that, the simple fact of possessing a Congolese passport that the applicant refused for a very specific reason to return to the Congolese authorities when requested by them to do so, constitutes proof of his intention to reavail himself of the protection of his country of nationality. The Court cannot accept such implied finding in the present affair in view of the inexistence of any credibility finding in the decision with respect to the applicant's explanations. 
     
      [24] 
      For the foregoing reasons the Court finds the Board's decision to be unreasonable. 

      [25] 
      The Court agrees with the parties that there is no question of general interest to certify... \\
    \bottomrule
    \end{tabularx}
    \caption{Example from COLIEE (Task 1.1) \cite{coliee2024}}
    \label{tab:ex_coliee}
\end{table*}

\begin{table*}[p]
    \centering
    \begin{tabularx}{\linewidth}{lX}
        \toprule
        Query & \small to constitute clear error. See United States v. Sullivan, 75 F.3d 297, 302-03 (7th Cir.1996). Throughout his briefs, Siegler attempts to portray the August 31 letter as
         a solicitation rather than a threat, in effect trying to challenge his conviction f
        or violating 18 U.S.C. § 876. By pleading guilty, however, Siegler admitted both of 
        the elements of Count II (mailing a threatening communication). See McCarthy v. Unit
        ed States, 394 U.S. 459, 466, 89 S.Ct. 1166, 22 L.Ed.2d 418 (1969) (“[A] guilty plea
         is an admission of all the elements of a formal criminal charge.”); United States v
        . Gilliam, 255 F.3d 428, 433 (7th Cir.2001) (same). In the written plea agreement an
        d during the plea hearing, Siegler admitted that on August 31, 1999, he wrote and ma
        iled to Hester a letter threatening Hauger; no more was required for a conviction un
        der 18 U.S.C. § 876. See  REDACTED .C. § 876 requires proof of two elements: (1) a t
        hreatening communication (2) was sent through the mail); United States v. Khorrami, 
        895 F.2d 1186, 1192 (7th Cir.1990) (conviction under 18 U.S.C. § 876 does not requir
        e proof that defendant intended to carry out threat). By admitting that the letter h
        e sent contained a threat within the meaning of 18 U.S.C. § 876, Siegler waived any 
        subsequent argument about the nature of the threat. See United States v. Newman, 148
         F.3d 871, 876 (7th Cir.1998) (defendant’s stipulation to conduct in plea agreement 
        conclusively admitted facts and waived subsequent challenge to them). Accordingly, S
        iegler’s argument that the letter did not contain a “true threat” is irrelevant to h
        is appeal of his sentence. Siegler also argues that because he did not send the lett
        er to Hauger or directly communicate the threat to her, there   \\
         Gold Passage(s) & \small FLAUM, Circuit Judge. For a period of more than four years, Richard Geisler was involved in a romantic relationship with Tena Camille DeAck
        len. During this time, the couple shared a joint bank account. Their relationship ended in early 1992, and Geisler thereafter contended tha
        t DeAcklen improperly withdrew \$1,280 of his money from their joint account. DeAck-len refused to repay this money, whereupon Geisler — who
         is white — began sending racially-charged, threatening letters to DeAck-len — who is African-American. In the end, Geisler sent six of the
        se hateful letters between September 1994 and January 1996. The district court convicted Geisler of six counts of mailing threatening commu
        nications with the intent to extort money in violation of 18 U.S.C. § 876. We affirm his convictions. Geisler stipulated at trial that he a
        uthored the letters that formed the basis for the charged offenses. There was similarly no dispute that he had sent the letters through the
         mails. Finally, Geisler did not— nor could he — challenge that the threats of injury and death (along with references to his “friends” aff
        iliated with the Ku Klux Klan who might assist him in carrying out these threats) contained in these letters constituted threats sufficient
         to trigger § 876. Rather, his challenge on appeal focuses on the fact that DeAcklen did not read all of the threatening letters that he se
        nt through the mails. Indeed, she testified that she read one letter in January 1995, as well as one or two others (she could not remember
        precisely), but that she turned over the other letters directly to the FBI without opening them. Geisler contends that, because DeAcklen ne
        ver received the threats contained in some of his letters, he did not violate § 876 on those counts. This argument reflects a patently inco
        rrect interpretation of,, the requirements of § 876 and our Circuit's precedent, and Geisler recognizes as much. Under thé plain language o
        f the statute, the Government only needed to prove that Geisler sent a communication through the mails that contained a threat to injure De
        Acklen; Geisler’s proposed “receipt” requirement is nowhere to be found in the statute. For this reason, we have stated repeatedly that the
         only two elements of a § 876 violation are (1) a threatening communication (2) sent through the mails. See, e.g., United States v. Sulliva
        n, 75 F.3d 297, 302 (7th Cir.1996) (“The sending of threatening communications is a crime quite apart from any intent to carry out the thre
        ats.”); United States v. Aman, 31 F.3d 550, 551 (7th Cir.1994) (stating that § 876 “prohibits the mailing of threatening communications”);
        United States v. Johnson, 965 F.2d 460, 467 (7th Cir.1992) (noting that § 876 “simply re-quirfes] that a defendant knowingly cause to be de
        livered a threatening letter in the U.S. mails”). In light of the plain language of the statute, it is not surprising that other Circuits s
        hare our view that there are only two required elements of a § 876 violation. See, e.g., United States v. Turner, 960 F.2d 461, 463 n. 2 (5
        th Cir.1992); United States v. Davis, 926 F.2d 969, 971 (10th Cir.), cert. denied, 500 U.S. 926, 111 S.Ct. 2036, 114 L.Ed.2d 121 (1991); Un
        ited States v. Davis, 876 F.2d 71, 73 (9th Cir.), cert. denied, 493 U.S. 866, 110 S.Ct. 188, 107 L.Ed.2d 143 (1989); United States v. Linco
        ln, 589 F.2d 379, 381 (8th Cir.1979); United States v. Chatman, 584 F.2d 1358, 1361 (4th Cir.1978). We reject Geisler’s attempt to create a
         new element of the offense... \\
    \bottomrule
    \end{tabularx}
    \caption{Example from CLERC \cite{hou2024clerc}}
    \label{tab:ex_clerc}
\end{table*}

\section{Query Expansion Prompts}
\label{sec:query_expansion_prompts}
Table \ref{tab:query_expansion_prompts_barexamqa} shows the query expansion prompts for Bar Exam QA. Table \ref{tab:query_expansion_prompts_housingqa} shows the query expansion prompts for Housing Statute QA.

\begin{table*}[p]
    \centering
    \begin{tabularx}{\linewidth}{lX}
         \toprule
         Paraphrasing & Given a legal question, paraphrase the question in ``Paraphrase:''. \\
         \midrule
         Chain-of-Thought & Given a legal question in ``Question:'', answer the question in ``Answer:''. Explain your reasoning in ``Explanation:''. Think step by step.\\
         \midrule
         Structured Reasoning & Given a set of facts about a legal scenario in ``Question:'', identify the key legal issue that arises from the facts and provide the applicable legal rule in ``Rule:''. \\
         \bottomrule
    \end{tabularx}
    \caption{Query expansion prompts for Bar Exam QA query expansion}
    \label{tab:query_expansion_prompts_barexamqa}
\end{table*}

\begin{table*}[p]
    \centering
    \begin{tabularx}{\linewidth}{lX}
         \toprule
         Chain-of-Thought & Consider the housing statute for \{jurisdiction\} in the year 2021. Given a legal question in ``Question:'', answer the question in ``Answer:''. Explain your reasoning in "Explanation:". Think step by step. \\
         \midrule
         Structured Reasoning &  Consider the housing statute for \{jurisdiction\} in the year 2021. The question given in ``Question:'' is a legal question about housing and eviction law in \{jurisdiction\}. Provide the applicable legal rule in ``Rule:''. If you do not know the state law, provide governing rules that address the question under typical eviction law.\\
         \bottomrule
    \end{tabularx}
    \caption{Query expansion prompts for Housing Statute QA}
    \label{tab:query_expansion_prompts_housingqa}
\end{table*}

\section{Structured Reasoning Prompt Query Expansion Examples}
\label{sec:structured_reasoning_prompt_encodes}
For the question from Bar Exam QA in Table \ref{tab:question_for_prompt_expansion_comparison}, we provide the query expansions with the paraphrasing prompt, chain-of-thought prompt, and structured reasoning prompt, and the gold passage for comparison in Table \ref{tab:structured_reasoning_prompt_encodes}, to illustrate how the structured reasoning prompt expansion encodes implicit retrieval task steps and captures latent legal issues.

\begin{table*}[p]
    \centering
    \begin{tabularx}{\linewidth}{lX}
         \toprule
         Question & Pauline, an unmarried female, was prominent in the women's liberation movement. She recently gave birth to a baby and publicly announced that she had no intention of marrying the father or disclosing his identity. The local newspaper, Journal, decided to do a series of articles on Pauline, entitled "The Perils of Pauline." The first article about Pauline discussed her parents. The article correctly stated that Mary, her mother, had died recently and Frank, her father, is still living. The article referred to the fact that at the time of Pauline's birth there were rumors that she had been born six months after the marriage of Mary and Frank, that Frank was not in fact her father, and that a person identified as Albert, who had played minor roles in two motion pictures, was her real father. Albert has lived in retirement for the last ten years. If Pauline asserts a claim based on invasion of privacy against Journal for the statements in the first article about her birth and it is established that the statements are true, the most likely result is that Pauline will \\
         \bottomrule
    \end{tabularx}
    \caption{Example question from Bar Exam QA for prompt expansion comparison in Table \ref{tab:structured_reasoning_prompt_encodes}.}
    \label{tab:question_for_prompt_expansion_comparison}
\end{table*}

\begin{table*}[p]
    \centering
    \begin{tabularx}{\linewidth}{lX}
         \toprule
         Paraphrasing Expansion &  \small If Pauline files a lawsuit against the newspaper Journal for invasion of privacy regarding the information published about her birth, and it is proven that the statements made in the article are accurate, what is the most probable outcome for Pauline? \\
         \midrule
         Chain-of-Thought Expansion & \small Pauline will not succeed in her claim based on invasion of privacy against Journal. In order to establish a claim for invasion of privacy, Pauline would need to show that Journal's statements about her birth were false and that they invaded her reasonable expectation of privacy. However, in this case, it is mentioned that the statements made by Journal in the first article about Pauline's birth are true. It is established that there were rumors about the circumstances of her birth, including the identity of her father. Since the statements are true, Pauline cannot argue that Journal invaded her privacy by publishing this information. The truth of the statements serves as a valid defense for Journal, making it unlikely for Pauline to succeed in her claim. \\
         \midrule
         Structured Reasoning Expansion & \small The applicable legal rule in this case would likely be related to the tort of invasion of privacy. In the United States, the specific elements and standards for invasion of privacy can vary depending on the jurisdiction. However, a common framework for invasion of privacy claims involves four distinct categories:
         
         1. Intrusion upon seclusion: This refers to the unauthorized invasion into a person's private affairs or physical space in a way that would be highly offensive to a reasonable person.
         
         2. Public disclosure of private facts: This involves the public dissemination of private and confidential information about an individual that would be highly offensive to a reasonable person and is not of legitimate public concern.

         3. False light: This occurs when false or misleading information is publicly attributed to an individual, portraying them in a highly offensive and false manner.

         4. Appropriation of name or likeness: This refers to the unauthorized use of a person's name, likeness, or identity for commercial purposes, without their consent.

         Based on the facts provided, it seems that the issue relevant to Pauline's claim against the Journal would fall under the category of ""public disclosure of private facts."" If the statements made in the first article about Pauline's birth are true (i.e., that there were rumors regarding her parentage), it may be more difficult for Pauline to establish a claim for invasion of privacy. The truth of the statements could potentially be a defense for the Journal, as disclosing true information may carry more weight in the context of a public figure like Pauline. \\
         \midrule
         Gold Passage & \small There are a number of defenses to a public disclosure claim. Below, you'll find explanations of some of the most common defenses:
         
         Legitimate public interest:
         Whether the public has a legitimate interest in the facts-at-issue is a question that depends on the context of the case, and one in which there is no particular formula for the courts to follow. Whether this defense can be effectively asserted will depend largely on whether the person involved has made him or herself - in a temporary newsworthy capacity or a more permanent celebrity capacity - something of a public figure. In such cases, details of their private lives are more likely to be considered items of legitimate public interest. The passage of time may lessen the public interest in a given fact (the newsworthiness of it), which may weaken this defense.
         
         Consent:
         Consent is a total defense. If the plaintiff has consented in some way to the disclosure, whether through a release form or through accepting an interview, then he or she cannot pursue a claim for public disclosure of private fact.
         
         Public Record:
         Matters of public record, such as birth date, military service records, and others, are exempted. The defendant may claim this defense by showing that the disclosed fact was actually a matter of public record. However, it should be noted that, unlike defamation actions, truth is no defense to a claim for public disclosure of private facts. This means that a defendant cannot refute a claim by showing that the disclosed fact was actually true or accurate. \\
         \bottomrule
    \end{tabularx}
    \caption{Example of query expansions with different prompting methods and gold passage for the question from Bar Exam QA in Table \ref{tab:question_for_prompt_expansion_comparison}. The structured reasoning expansion most clearly identifies the legal issue and statement of the applicable legal rule that resembles the gold passage.}
    \label{tab:structured_reasoning_prompt_encodes}
\end{table*}

\section{Lexical Similarity Statistical Test Results}
\label{sec:lex_sim_stat_test_results}
In Tables \ref{tab:ks_test_qgp}, \ref{tab:ks_test_gpa}, \ref{tab:t_test_qgp}, \ref{tab:t_test_gpa}, we report the statistical test results comparing the lexical similarity distribution over Bar Exam QA and Housing Statute QA to other popular general and legal domain retrieval tasks.

\begin{table*}[p]
    \centering
    \begin{tabular}{llrr}
         \toprule
         Distribution 1 &  Distribution 2 & Test statistic (\emph{D}) & $p$-value \\
         \midrule
         Bar Exam QA & NQ & 0.590 & <0.001 \\
         Bar Exam QA & HotpotQA & 0.610 & <0.001 \\
         Bar Exam QA & COLIEE (Task 1.1) & 0.607 & <0.001 \\
         Bar Exam QA & CLERC & 0.630 & <0.001 \\
         \midrule
         Housing Statute QA & NQ & 0.596 & <0.001 \\
         Housing Statute QA & HotpotQA & 0.593 & <0.001 \\
         Housing Statute QA & COLIEE (Task 1.1) & 0.585 & <0.001 \\
         Housing Statute QA & CLERC & 0.614 & <0.001 \\
         \bottomrule
    \end{tabular}
    \caption{Kolmogorov-Smirnov test results comparing the lexical similarity (query, gold passage) distribution of Bar Exam QA and Housing Statute QA to the distribution of other general and legal domain IR tasks.}
    \label{tab:ks_test_qgp}
\end{table*}

\begin{table*}[p]
    \centering
    \begin{tabular}{llrr}
         \toprule
         Distribution 1 &  Distribution 2 & Test statistic (\emph{D}) & $p$-value \\
         \midrule
         Bar Exam QA & NQ & 0.251 & <0.001 \\
         Bar Exam QA & HotpotQA & 0.249 & <0.001 \\
         \midrule
         Housing Statute QA & NQ & 0.770 & <0.001 \\
         Housing Statute QA & HotpotQA & 0.771 & <0.001 \\
         \bottomrule
    \end{tabular}
    \caption{Kolmogorov-Smirnov test results comparing the lexical similarity (gold passage, answer) distribution of Bar Exam QA and Housing Statute QA to the distribution of other general domain IR tasks.}
    \label{tab:ks_test_gpa}
\end{table*}

\begin{table*}[p]
    \centering
    \begin{tabular}{llrr}
         \toprule
         Distribution 1 &  Distribution 2 & Test statistic (\emph{t}) & $p$-value \\
         \midrule
         Bar Exam QA & NQ & -58.1 & <0.001 \\
         Bar Exam QA & HotpotQA & -63.8 & <0.001 \\
         Bar Exam QA & COLIEE (Task 1.1) & -60.4 & <0.001 \\
         Bar Exam QA & CLERC & -60.1 & <0.001 \\
         \midrule
         Housing Statute QA & NQ & -79.4 & <0.001 \\
         Housing Statute QA & HotpotQA & -87.7 & <0.001 \\
         Housing Statute QA & COLIEE (Task 1.1) & -83.6 & <0.001 \\
         Housing Statute QA & CLERC & -80.5 & <0.001 \\
         \bottomrule
    \end{tabular}
    \caption{\emph{t}-test results comparing the mean lexical similarity (query, gold passage) of Bar Exam QA and Housing Statute QA to the mean lexical similarity of other general and legal domain IR tasks.}
    \label{tab:t_test_qgp}
\end{table*}

\begin{table*}[p]
    \centering
    \begin{tabular}{llrr}
         \toprule
         Distribution 1 &  Distribution 2 & Test statistic (\emph{t}) & $p$-value \\
         \midrule
         Bar Exam QA & NQ & -23.2 & <0.001 \\
         Bar Exam QA & HotpotQA & -24.1 & <0.001 \\
         \midrule
         Housing Statute QA & NQ & -99.0 & <0.001 \\
         Housing Statute QA & HotpotQA & -102 & <0.001 \\
         \bottomrule
    \end{tabular}
    \caption{\emph{t}-test results comparing the mean lexical similarity (gold passage, answer) distributions of Bar Exam QA and Housing Statute QA to the mean lexical similarity of other general domain IR tasks.}
    \label{tab:t_test_gpa}
\end{table*}

\section{Retrieval Results}
\label{sec:comp_retrieval_results}
For Bar Exam QA, Table \ref{tab:retrieval_perf_mbe}, \ref{tab:retrieval_perf_barexamqa-mbe}, and \ref{tab:retrieval_perf_barexamqa-barbri} report full retrieval performance evaluation results for the aggregate, historical MBE, and Barbri subsets. 

For Housing Statute QA, Table \ref{tab:retrieval_perf_housingqa_upper} reports full retrieval performance evaluation results. Since we transform the original questions to binary classification (Y/N) questions for the downstream task, some gold statute passages are relevant for the original question, but not the derived Y/N question. In Table \ref{tab:retrieval_perf_housingqa_upper}, recall is computed as the retrieval of at least one gold passage (upper bound). For completeness, we also provide Table \ref{tab:retrieval_perf_housingqa_lower}, where recall is computed as the retrieval of all gold passages (lower bound); as an example has a maximum of 10 gold passages, we report this value from retrieval depth of $k = 10$.

For NQ, Table \ref{tab:retrieval_perf_nq} reports retrieval performance evaluation results.

For HotpotQA, Table \ref{tab:retrieval_perf_hotpotqa} reports retrieval performance evaluation results. We compute recall as the retrieval of the two gold passages, since both gold passages are required to answer each question; we report this value from retrieval depth of $k = 2$.

For COLIEE, Table \ref{tab:retrieval_perf_coliee} reports retrieval performance results on Task 1.1 For CLERC, Table \ref{tab:retrieval_perf_clerc} reports retrieval performance results for the document setting.

\begin{table*}[p]
    \centering
    \begin{tabular}{lrrrrr}
         \toprule
         Method & Recall@1 & Recall@10 & MRR@10 & Recall@100 & Recall@1000 \\
         \midrule
         BM25 & \\
         \hspace{5mm} Baseline & 1.83 & 5.03 & 2.68 & 9.99 & 20.52 \\
         \hspace{5mm} Paraphrase & 1.77 & 5.03 & 2.77 & 10.60 & 22.45 \\
         \hspace{5mm} CoT & 3.70 & 9.65 & 5.28 & 22.18 & 42.76 \\
         \hspace{5mm} Structured reasoning & 3.43 & 11.31 & 5.51 & 26.39 & 47.86 \\
         $\text{E5}_{\text{small-v2}}$ & \\
         \hspace{5mm} Baseline & 2.51 & 5.30 & 3.39 & 11.38 & 21.88 \\
         \hspace{5mm} Paraphrase & 1.90 & 5.26 & 2.88 & 10.50 & 21.26 \\
         \hspace{5mm} CoT & 3.09 & 8.93 & 4.72 & 18.48 & 33.56 \\
         \hspace{5mm} Structured reasoning & 3.53 & 10.02 & 5.25 & 23.03 & 38.25 \\
         $\text{E5}_{\text{base-v2}}$ & \\
         \hspace{5mm} Baseline & 3.33 & 8.42 & 4.71 & 16.34 & 29.14 \\
         \hspace{5mm} Paraphrase & 2.89 & 7.74 & 4.17 &  15.15 & 29.69 \\
         \hspace{5mm} CoT & 5.88 & 15.79 & 8.55 & 31.32 & 49.01 \\
         \hspace{5mm} Structured reasoning & 6.25 & 17.60 & 9.52 & 33.42 & 50.34 \\
         $\text{E5}_{\text{large-v2}}$ & \\
         \hspace{5mm} Baseline & 3.13 & 7.00 & 4.25 & 15.35 & 27.00 \\
         \hspace{5mm} Paraphrase & 2.48 & 6.11 & 3.46 & 13.49 & 24.49 \\
         \hspace{5mm} CoT & 4.82 & 12.77 & 6.91 & 26.80 & 43.75 \\
         \hspace{5mm} Structured reasoning & 5.84 & 15.86 & 8.69 & 32.34 & 49.01 \\
         $\text{E5}_{\text{mistral-7b}}$ & \\
         \hspace{5mm} Baseline & 5.71 & 15.25 & 8.19 & 34.10 & 56.28 \\
         \hspace{5mm} Paraphrase & 4.96 & 13.08 & 7.15 & 31.32 & 53.06 \\
         \hspace{5mm} CoT & 3.46 & 13.32 & 5.88 & 35.29 & 58.51 \\
         \hspace{5mm} Structured reasoning & 2.81 & 14.03 & 5.64 & 37.40 & 60.73 \\
         \bottomrule
    \end{tabular}
    \caption{Retrieval performance on Bar Exam QA, aggregated.} 
    \label{tab:retrieval_perf_mbe}
\end{table*}

\begin{table*}[p]
    \centering
    \begin{tabular}{lrrrrr}
         \toprule
         Method & Recall@1 & Recall@10 & MRR@10 & Recall@100 & Recall@1000 \\
         \midrule
         BM25 & \\ 
        \hspace{5mm} Baseline & 0.25 & 0.75 & 0.37 & 2.26 & 8.79 \\
        \hspace{5mm} Paraphrase & 0.33 & 0.84 & 0.45 & 2.85 & 10.71 \\
        \hspace{5mm} CoT & 0.75 & 2.68 & 1.2 & 8.87 & 27.53 \\
        \hspace{5mm} Structured reasoning & 0.59 & 3.1 & 1.21 & 12.55 & 32.05 \\
        $\text{E5}_{\text{small-v2}}$ & \\
        \hspace{5mm} Baseline & 0.08 & 0.59 & 0.18 & 2.68 & 9.29 \\
        \hspace{5mm} Paraphrase & 0.08 & 0.92 & 0.27 & 3.26 & 9.54 \\
        \hspace{5mm} CoT & 0.25 & 2.34 & 0.81 & 8.95 & 20.5 \\
        \hspace{5mm} Structured reasoning & 0.42 & 2.34 & 0.94 & 10.96 & 23.6 \\
        $\text{E5}_{\text{base-v2}}$ & \\
        \hspace{5mm} Baseline & 0.25 & 0.84 & 0.39 & 3.51 & 11.21 \\
        \hspace{5mm} Paraphrase & 0.17 & 0.75 & 0.35 & 3.85 & 14.39 \\
        \hspace{5mm} CoT & 0.75 & 4.1 & 1.55 & 13.56 & 27.7 \\
        \hspace{5mm} Structured reasoning & 1.0 & 4.1 & 1.74 & 12.8 & 29.37 \\
        $\text{E5}_{\text{large-v2}}$ & \\
        \hspace{5mm} Baseline & 0.17 & 0.92 & 0.34 & 4.27 & 12.3 \\
        \hspace{5mm} Paraphrase & 0.08 & 0.84 & 0.23 & 3.6 & 10.88 \\
        \hspace{5mm} CoT & 0.67 & 4.18 & 1.56 & 14.23 & 28.7 \\
        \hspace{5mm} Structured reasoning & 1.34 & 5.19 & 2.42 & 16.49 & 31.8 \\
        $\text{E5}_{\text{mistral-7b}}$ & \\
        \hspace{5mm} Baseline & 0.84 & 3.26 & 1.45 & 9.71 & 26.36 \\
        \hspace{5mm} Paraphrase & 0.5 & 1.76 & 0.84 & 7.11 & 18.41 \\
        \hspace{5mm} CoT & 1.26 & 5.86 & 2.39 & 20.33 & 42.43 \\
        \hspace{5mm} Structured reasoning & 0.67 & 6.95 & 2.26 & 23.51 & 42.85 \\
         \bottomrule
    \end{tabular}
    \caption{Retrieval performance on Bar Exam QA, disaggregated (Historical MBE subset).} 
    \label{tab:retrieval_perf_barexamqa-mbe}
\end{table*}

\begin{table*}[p]
    \centering
    \begin{tabular}{lrrrrr}
         \toprule
         Method & Recall@1 & Recall@10 & MRR@10 & Recall@100 & Recall@1000 \\
         \midrule
         BM25 & \\
        \hspace{5mm} Baseline & 2.81 & 7.66 & 4.11 & 14.71 & 27.49 \\
        \hspace{5mm} Paraphrase & 2.75 & 7.05 & 3.98 & 14.6 & 28.21 \\
        \hspace{5mm} CoT & 5.51 & 13.88 & 7.77 & 30.14 & 51.24 \\
        \hspace{5mm} Structured reasoning & 5.18 & 16.31 & 8.14 & 34.55 & 56.53 \\
         $\text{E5}_{\text{small-v2}}$ & \\
        \hspace{5mm} Baseline & 3.09 & 7.66 & 4.43 & 14.55 & 27.38 \\
        \hspace{5mm} Paraphrase & 2.81 & 6.94 & 4.05 & 14.27 & 28.48 \\
        \hspace{5mm} CoT & 4.35 & 12.12 & 6.54 & 23.2 & 40.22 \\
        \hspace{5mm} Structured reasoning & 4.9 & 12.89 & 7.06 & 28.21 & 44.79 \\
        $\text{E5}_{\text{base-v2}}$ & \\
        \hspace{5mm} Baseline & 3.86 & 10.3 & 5.71 & 20.17 & 37.47 \\
        \hspace{5mm} Paraphrase & 3.31 & 7.99 & 4.64 & 17.47 & 32.45 \\
        \hspace{5mm} CoT & 7.49 & 18.62 & 10.49 & 38.07 & 58.46 \\
        \hspace{5mm} Structured reasoning & 8.1 & 23.25 & 12.52 & 44.35 & 62.87 \\
        $\text{E5}_{\text{large-v2}}$ & \\
        \hspace{5mm} Baseline & 4.52 & 10.03 & 6.13 & 20.61 & 34.99 \\
        \hspace{5mm} Paraphrase & 3.42 & 7.49 & 4.52 & 15.65 & 29.2 \\
        \hspace{5mm} CoT & 6.78 & 17.63 & 9.54 & 33.5 & 51.85 \\
        \hspace{5mm} Structured reasoning & 7.16 & 19.83 & 10.72 & 39.23 & 57.13 \\
        $\text{E5}_{\text{mistral-7b}}$ & \\
        \hspace{5mm} Baseline & 5.29 & 11.85 & 7.05 & 25.79 & 41.82 \\
        \hspace{5mm} Paraphrase & 2.09 & 5.51 & 3.05 & 13.44 & 26.61 \\
        \hspace{5mm} CoT & 3.14 & 11.85 & 5.38 & 27.55 & 47.93 \\
        \hspace{5mm} Structured reasoning & 1.93 & 10.96 & 4.2 & 28.1 & 48.54 \\
         \bottomrule
    \end{tabular}
    \caption{Retrieval performance on Bar Exam QA, disaggregated (Barbri subset).} 
    \label{tab:retrieval_perf_barexamqa-barbri}
\end{table*}

\begin{table*}[p]
    \centering
    \begin{tabular}{lrrrrr}
         \toprule
         Method & Recall@1 & Recall@10 & MRR@10 & Recall@100 & Recall@1000 \\
         \midrule
         BM25 & \\
         \hspace{5mm} Baseline & 14.72 & 40.81 & 21.99 & 62.41 & 76.19 \\
         \hspace{5mm} CoT & 14.07 & 43.43 & 22.11 & 64.22 & 77.0 \\
         \hspace{5mm} Structured reasoning & 18.68 & 51.09 & 27.74 & 71.76 & 81.69 \\
         $\text{E5}_{\text{small-v2}}$ & \\
        \hspace{5mm} Baseline & 8.97 & 34.35 & 15.95 & 64.42 & 81.63 \\
        \hspace{5mm} CoT & 8.86 & 33.71 & 15.7 & 57.9 & 76.62 \\
        \hspace{5mm} Structured reasoning & 13.38 & 42.0 & 21.34 & 69.02 & 83.41 \\
         $\text{E5}_{\text{base-v2}}$ & \\
         \hspace{5mm} Baseline & 13.56 & 45.75 & 22.38 & 74.61 & 86.65 \\
        \hspace{5mm} CoT & 13.24 & 40.83 & 21.19 & 68.89 & 84.63 \\
        \hspace{5mm} Structured reasoning & 16.56 & 51.36 & 26.45 & 76.21 & 87.25 \\
         $\text{E5}_{\text{large-v2}}$ & \\
         \hspace{5mm} Baseline & 16.08 & 50.58 & 26.02 & 78.83 & 87.73 \\
        \hspace{5mm} CoT & 12.91 & 43.76 & 21.61 & 70.67 & 84.9 \\
        \hspace{5mm} Structured reasoning & 17.86 & 52.74 & 28.01 & 78.11 & 87.48 \\
         $\text{E5}_{\text{mistral-7b}}$ & \\
         \hspace{5mm} Baseline & 25.36 & 65.31 & 37.7 & 84.31 & 88.87 \\
        \hspace{5mm} CoT & 26.25 & 64.54 & 38.13 & 82.58 & 88.68 \\
        \hspace{5mm} Structured reasoning & 30.21 & 68.79 & 42.54 & 85.04 & 89.1 \\
         \bottomrule
    \end{tabular}
    \caption{Retrieval performance on Housing Statute QA. Recall is computed as retrieval of at least one gold passage for a given query (upper bound).}
    \label{tab:retrieval_perf_housingqa_upper}
\end{table*}

\begin{table*}[p]
    \centering
    \begin{tabular}{lrrrr}
         \toprule
         Method & Recall@10 & MRR@10 & Recall@100 & Recall@1000 \\
         \midrule
         BM25 & \\
         \hspace{5mm} Baseline & 18.31 & 10.35 & 39.89 & 59.68 \\
        \hspace{5mm} CoT & 18.85 & 9.81 & 42.04 & 60.44 \\
        \hspace{5mm} Structured reasoning & 23.83 & 13.73 & 49.53 & 65.1 \\
         $\text{E5}_{\text{small-v2}}$ & \\
         \hspace{5mm} Baseline & 13.32 & 6.53 & 37.57 & 66.32 \\
        \hspace{5mm} CoT & 13.63 & 7.12 & 32.23 & 58.66 \\
        \hspace{5mm} Structured reasoning & 18.12 & 9.35 & 43.11 & 68.52 \\
         $\text{E5}_{\text{base-v2}}$ & \\
         \hspace{5mm} Baseline & 20.3 & 10.4 & 51.61 & 75.53 \\
        \hspace{5mm} CoT & 17.57 & 9.54 & 44.89 & 70.22 \\
        \hspace{5mm} Structured reasoning & 23.97 & 12.25 & 53.67 & 75.75 \\
         $\text{E5}_{\text{large-v2}}$ & \\
         \hspace{5mm} Baseline & 24.4 & 13.22 & 57.2 & 78.36 \\
        \hspace{5mm} CoT & 20.28 & 10.22 & 45.05 & 70.06 \\
        \hspace{5mm} Structured reasoning & 26.0 & 13.68 & 54.44 & 77.0 \\
         $\text{E5}_{\text{mistral-7b}}$ & \\
         \hspace{5mm} Baseline & 35.77 & 21.22 & 68.96 & 81.02 \\
        \hspace{5mm} CoT & 35.4 & 21.11 & 65.77 & 79.79 \\
        \hspace{5mm} Structured reasoning & 39.33 & 24.72 & 69.18 & 81.12 \\
         \bottomrule
    \end{tabular}
    \caption{Retrieval performance on Housing Statute QA. Recall is computed as retrieval of all gold passages for a given query (lower bound). Since a query has a maximum of 10 potentially relevant gold passages, this metric is computed for $k \geq 10$.}
    \label{tab:retrieval_perf_housingqa_lower}
\end{table*}

\begin{table*}[p]
    \centering
    \begin{tabular}{lrrrrr}
         \toprule
         Method & Recall@1 & Recall@10 & MRR@10 & Recall@100 & Recall@1000 \\
         \midrule
         BM25 & \\
         \hspace{5mm} Baseline & 13.04 & 40.44 & 21.13 & 67.06 & 81.69 \\
        \hspace{5mm} CoT & 22.83 & 57.44 & 33.62 & 79.87 & 89.77 \\
         $\text{E5}_{\text{small-v2}}$ & \\
         \hspace{5mm} Baseline & 27.38 & 64.54 & 39.12 & 85.14 & 94.5 \\
        \hspace{5mm} CoT & 31.89 & 68.37 & 43.65 & 86.96 & 94.24 \\
        $\text{E5}_{\text{base-v2}}$ & \\
        \hspace{5mm} Baseline & 29.11 & 66.66 & 40.73 & 86.85 & 95.37 \\
        \hspace{5mm} CoT & 31.4 & 68.25 & 43.04 & 87.46 & 95.22 \\
        $\text{E5}_{\text{large-v2}}$ & \\
        \hspace{5mm} Baseline & 30.45 & 68.68 & 42.59 & 88.62 & 95.97 \\
        \hspace{5mm} CoT & 32.73 & 70.45 & 44.8 & 88.59 & 95.71 \\
        $\text{E5}_{\text{mistral-7b}}$ & \\
        \hspace{5mm} Baseline & 3.36 & 11.15 & 5.49 & 22.33 & 38.21 \\
        \hspace{5mm} CoT & 28.13 & 58.14 & 37.57 & 76.36 & 87.8 \\
        \bottomrule
    \end{tabular}
    \caption{Retrieval performance on Natural Questions.} 
    \label{tab:retrieval_perf_nq}
\end{table*}

\begin{table*}[p]
    \centering
    \begin{tabular}{lrrrrr}
         \toprule
         Method & Recall@2 & Recall@10 & MRR@10 & Recall@100 & Recall@1000 \\
         \midrule
         BM25 & \\
         \hspace{5mm} Baseline & 13.1 & 32.72 & 29.26 & 56.89 & 74.73 \\
        \hspace{5mm} CoT & 14.41 & 37.03 & 32.72 & 58.69 & 74.41 \\
         $\text{E5}_{\text{small-v2}}$ & \\
         \hspace{5mm} Baseline & 19.27 & 45.0 & 40.52 & 66.47 & 81.62 \\
        \hspace{5mm} CoT & 21.36 & 46.97 & 42.41 & 65.29 & 79.26 \\
        $\text{E5}_{\text{base-v2}}$ & \\
        \hspace{5mm} Baseline & 23.28 & 49.55 & 45.36 & 70.68 & 86.01 \\
        \hspace{5mm} CoT & 21.35 & 46.75 & 42.3 & 65.24 & 79.7 \\
        $\text{E5}_{\text{large-v2}}$ & \\
        \hspace{5mm} Baseline & 27.1 & 56.18 & 51.73 & 75.75 & 89.01 \\
        \hspace{5mm} CoT & 23.98 & 50.18 & 45.57 & 68.32 & 81.0 \\
        $\text{E5}_{\text{mistral-7b}}$ & \\
        \hspace{5mm} Baseline & 5.55 & 17.25 & 14.68 & 35.58 & 57.47 \\
        \hspace{5mm} CoT & 17.57 & 39.73 & 35.72 & 58.42 & 74.58 \\
        \bottomrule
    \end{tabular}
    \caption{Retrieval performance on HotpotQA.} 
    \label{tab:retrieval_perf_hotpotqa}
\end{table*}

\begin{table*}[p]
    \centering
    \begin{tabular}{lrrrrr}
         \toprule
         Method & Recall@1 & Recall@10 & MRR@10 & Recall@100 & Recall@1000 \\
         \midrule
         BM25 & \\
         \hspace{5mm} Baseline & 0.0 & 38.11 & 10.92 & 71.6 & 92.33 \\
         \hspace{5mm} CoT & 0.08 & 40.77 & 11.56 & 75.27 & 94.76 \\
         $\text{E5}_{\text{small-v2}}$ & \\
         \hspace{5mm} Baseline & 0.0 & 27.15 & 8.53 & 59.94 & 87.72 \\
         \hspace{5mm} CoT & 1.8 & 30.13 & 10.0 & 63.93 & 89.36 \\
        $\text{E5}_{\text{base-v2}}$ & \\
        \hspace{5mm} Baseline & 0.0 & 28.09 & 8.55 & 59.62 & 88.58 \\
        \hspace{5mm} CoT & 3.05 & 32.47 & 11.53 & 65.96 & 91.47 \\
        $\text{E5}_{\text{large-v2}}$ & \\
        \hspace{5mm} Baseline & 0.0 & 32.71 & 9.97 & 63.15 & 88.89 \\
        \hspace{5mm} CoT & 2.27 & 33.8 & 11.42 & 68.62 & 92.1 \\
        $\text{E5}_{\text{mistral-7b}}$ & \\
        \hspace{5mm} Baseline & 0.0 & 52.11 & 16.42 & 83.02 & 96.56 \\
        \hspace{5mm} CoT & 0.86 & 38.97 & 12.46 & 78.09 & 95.46 \\
        \bottomrule
    \end{tabular}
    \caption{Retrieval performance on COLIEE.} 
    \label{tab:retrieval_perf_coliee}
\end{table*}

\begin{table*}[p]
    \centering
    \begin{tabular}{lrrrrr}
         \toprule
         Method & Recall@1 & Recall@10 & MRR@10 & Recall@100 & Recall@1000 \\
         \midrule
         BM25 & \\
         \hspace{5mm} Baseline & 0.11 & 11.75 & 3.43 & 27.57 & 48.26 \\
         \hspace{5mm} CoT & 0.63 & 8.52 & 2.78 & 22.2 & 40.06 \\
         $\text{E5}_{\text{small-v2}}$ & \\
         \hspace{5mm} Baseline & 1.16 & 5.02 & 2.28 & 10.94 & 21.89 \\
         \hspace{5mm} CoT & 0.84 & 3.65 & 1.64 & 8.31 & 17.33 \\
        $\text{E5}_{\text{base-v2}}$ & \\
        \hspace{5mm} Baseline & 1.33 & 5.3 & 2.4 & 11.29 & 22.9 \\
        \hspace{5mm} CoT & 0.84 & 3.89 & 1.68 & 8.59 & 18.41 \\
        $\text{E5}_{\text{large-v2}}$ & \\
        \hspace{5mm} Baseline & 1.3 & 6.8 & 2.9 & 14.42 & 25.68 \\
        \hspace{5mm} CoT & 0.84 & 5.4 & 2.1 & 12.35 & 23.82 \\
        $\text{E5}_{\text{mistral-7b}}$ & \\
        \hspace{5mm} Baseline & 1.37 & 8.49 & 3.35 & 19.22 & 32.3 \\
        \hspace{5mm} CoT & 0.81 & 4.91 & 1.93 & 12.63 & 23.33 \\
        \bottomrule
    \end{tabular}
    \caption{Retrieval performance on CLERC.} 
    \label{tab:retrieval_perf_clerc}
\end{table*}

\section{Downstream QA Results}
\label{sec:downstream_qa_results}
Tables \ref{tab:qa_perf_barexamqa}, \ref{tab:qa_perf_barexamqa_large_model}, and \ref{tab:qa_perf_housingqa} show downstream QA results for Llama 3 8B Instruct and LLama 3 70B Instruct on Bar Exam QA and for LLama 3 8B Instruct on Housing Statute QA. Tables \ref{tab:qa_perf_barexamqa_gpt4omini} and \ref{tab:qa_perf_housingqa_gpt4omini} show downstream QA results for GPT-4o-mini (gpt-4o-mini-2024-07-18) on Bar Exam QA and Housing Statute QA.

\begin{table*}[p]
    \centering
    \begin{tabular}{lrr}
         \toprule
         Retrieval Method & Accuracy (Top 1) & Accuracy (Top 10) \\
         \midrule
         No passage & 37.84 & - \\
         BM25 & 39.10 & 41.14 \\
         BM25 + reasoning & 42.82 & 44.42 \\
         $\text{E5}_{\text{small-v2}}$ & 39.24 & 41.06 \\
         $\text{E5}_{\text{small-v2}}$ + reasoning & 41.69 & 44.72 \\
         $\text{E5}_{\text{base-v2}}$ & 39.67 & 41.99  \\
         $\text{E5}_{\text{base-v2}}$ + reasoning & 42.46 & 44.98 \\
         $\text{E5}_{\text{large-v2}}$ & 40.63 & 41.83 \\
         $\text{E5}_{\text{large-v2}}$ + reasoning & 42.09 & 44.75 \\
         $\text{E5}_{\text{mistral-7b}}$ & 41.33 & 43.52 \\
         $\text{E5}_{\text{mistral-7b}}$ + reasoning & 42.46 & 44.39 \\
         Reasoning rollout as pseudo-passage & 42.69 & \\
         Gold passage & 57.38 & - \\
         \bottomrule
    \end{tabular}
    \caption{Downstream task performance on Bar Exam QA on Llama-3-8B-Instruct. We perform coarse retrieval of the top $k$ passages per query using the retrieval method in the first column of the table. + reasoning indicates the structured reasoning query expansion method. Then, we rerank  the top $k$ retrieved passages using the predicted answer confidence from the Llama-3-8B-Instruct model. Accuracy (Top $k$) is calculated using the passage that gives the maximum confidence answer prediction from the top $k$ retrieved passages.} 
    \label{tab:qa_perf_barexamqa}
\end{table*}

\begin{table*}[p]
    \centering
    \begin{tabular}{lr}
         \toprule
         Retrieval Method & Accuracy \\
         \midrule
         No passage & 52.22 \\
         Gold passage & 68.80 \\
         \bottomrule
    \end{tabular}
    \caption{Downstream task performance on Bar Exam QA on Llama-3-70B-Instruct. We evaluate for no passage and gold passage to show that the gold passage quality is high, but the Llama-3-8B-Instruct struggles to apply them to the question.} 
    \label{tab:qa_perf_barexamqa_large_model}
\end{table*}

\begin{table*}[p]
    \centering
    \begin{tabular}{lr}
         \toprule
         Retrieval Method & Accuracy \\
         \midrule
         No passage & 49.73 \\
         Reasoning rollout as pseudo-passage & 47.94 \\
         Retrieved passage & 50.23 \\
         Gold passage & 62.76 \\
         \bottomrule
    \end{tabular}
    \caption{Downstream task performance on Bar Exam QA on GPT-4o-mini (gpt-4o-mini-2024-07-18). We evaluate for no passage, retrieved passage ($\text{E5}_{\text{mistral-7b}}$ + structured reasoning), the generative reasoning rollout from the structured reasoning query expansion method as a pseudo-passage, and the gold passage.}
    \label{tab:qa_perf_barexamqa_gpt4omini}
\end{table*}

\begin{table*}[p]
    \centering
    \begin{tabular}{lrr}
         \toprule
         Retrieval Method & Accuracy (Top 1) & Accuracy (Top 10) \\
         \midrule
         No passage & 58.81 & - \\
         BM25 & 68.22 & 66.43 \\
         BM25 + reasoning & 67.43 & 66.18 \\
         $\text{E5}_{\text{small-v2}}$ & 67.59 & 65.83 \\
         $\text{E5}_{\text{small-v2}}$ + reasoning & 67.30 & 65.65 \\
         $\text{E5}_{\text{base-v2}}$ & 68.10 & 66.14 \\
         $\text{E5}_{\text{base-v2}}$ + reasoning & 68.19 & 65.72 \\
         $\text{E5}_{\text{large-v2}}$ & 68.43 & 66.14 \\
         $\text{E5}_{\text{large-v2}}$ + reasoning & 67.95 & 65.72  \\
         $\text{E5}_{\text{mistral-7b}}$ & 70.26 & 66.81 \\
         $\text{E5}_{\text{mistral-7b}}$ + reasoning & 69.03 & 66.58 \\
         Reasining rollout as pseudo-passage & 70.23 & \\
         Gold passage & 75.27 & - \\
         \bottomrule
    \end{tabular}
    \caption{Downstream task performance on Housing Statute QA on Llama-3-8B-Instruct. We perform coarse retrieval of the top $k$ passages per query using the retrieval method in the first column of the table. + reasoning indicates the structured reasoning query expansion method. Then, we rerank  the top $k$ retrieved passages using the predicted answer confidence from the Llama-3-8B-Instruct model. Accuracy (Top $k$) is calculated using the passage that gives the maximum confidence answer prediction from the top $k$ retrieved passages.}
    \label{tab:qa_perf_housingqa}
\end{table*}

\begin{table*}[p]
    \centering
    \begin{tabular}{lr}
         \toprule
         Retrieval Method & Accuracy \\
         \midrule
         No passage & 48.18 \\
         Reasoning rollout as pseudo-passage & 68.51 \\
         Retrieved passage & 71.71 \\
         Gold passage & 77.98 \\
         \bottomrule
    \end{tabular}
    \caption{Downstream task performance on Housing Statute QA on GPT-4o-mini (gpt-4o-mini-2024-07-18). We evaluate for no passage, retrieved passage ($\text{E5}_{\text{mistral-7b}}$ + structured reasoning), the generative reasoning rollout from the structured reasoning query expansion method as a pseudo-passage, and the gold passage.} 
    \label{tab:qa_perf_housingqa_gpt4omini}
\end{table*}

\end{document}